\documentclass[lettersize,journal]{IEEEtran}
\usepackage{amsmath,amsfonts}
\usepackage{algorithmic}
\usepackage{algorithm}
\usepackage{array}
\usepackage{amsthm}
\usepackage{textcomp}
\usepackage{epsfig}
\usepackage{stfloats}
\usepackage{url}
\usepackage{verbatim}
\usepackage{graphicx}
\usepackage{caption}
\usepackage{subcaption}  
\usepackage{cite}
\usepackage{graphicx}
\usepackage{amsthm} 
\newtheorem{theorem}{Theorem}
\newtheorem{lemma}{Lemma}
\usepackage[justification=centering]{caption} 
\usepackage[utf8]{inputenc}
\usepackage{balance}
  
\def\BibTeX{{\rm B\kern-.05em{\sc i\kern-.025em b}\kern-.08em
    T\kern-.1667em\lower.7ex\hbox{E}\kern-.125emX}}
\begin{document}
\allowdisplaybreaks 
\title{MAB-Based Channel Scheduling for Asynchronous Federated Learning in Non-Stationary Environments}
\author{Zhiyin Li, Yubo Yang, Tao Yang,~\IEEEmembership{Member,~IEEE,} Ziyu Guo,~\IEEEmembership{Member,~IEEE,} Xiaofeng Wu, Bo Hu,~\IEEEmembership{Member,~IEEE}}


\maketitle
\begin{abstract}
Federated learning enables distributed model training across clients under central coordination without raw data exchange. However, in wireless implementations, frequent parameter updates between the server and clients create significant communication overhead. While existing research assumes either known channel state information (CSI) or that the channel follows a stationary distribution, practical wireless channels exhibit non-stationary characteristics due to channel fading, user mobility, and hostile attacks in telecommunication networks. The unavailability of both CSI and time-varying channel distribution can lead to unpredictable failures in parameter transmission, exacerbating clients staleness thus affecting model convergence. To address these challenges, we propose an asynchronous federated learning scheduling framework for non-stationary channel environments, designed to reduce clients staleness while promoting both fair and efficient communication and aggregation. This framework considers two channel scenarios: extremely non-stationary and piecewise-stationary channels. Age of Information (AoI) serves as a metric to quantify client staleness under non-stationary conditions. Firstly, we perform a rigorous convergence analysis to explore the impact of AoI and per-round client participation on learning performance. The channel scheduling problem in the non-stationary scenario is addressed and formulated within the multi-armed bandit (MAB) framework and we derive the achievable theoretical lower bounds on the AoI regret. Based on this framework, we propose corresponding scheduling strategies for the two non-stationary channel scenarios that leverage the foundations of the GLR-CUCB and M-exp3 algorithms, along with derivations of their respective upper bounds on AoI regret. Additionally, to address the issue of imbalanced client updates in non-stationary channels, we introduce an adaptive matching strategy that incorporates considerations of marginal utility and fairness of clients. Simulation results demonstrate that the proposed algorithm achieves sub-linear growth in AoI regret, accelerates federated learning convergence, and promotes fairer aggregation.

\end{abstract}

\section{Introduction}


\IEEEPARstart{T}{he} proliferation of Internet of Things (IoT) devices and the rise of edge computing have resulted in an increasingly decentralized distribution of data across end devices, such as smartphones and sensors. In traditional centralized machine learning approaches, data consolidation at a single location is required, which raises privacy concerns and incurs significant communication overhead. In contrast, federated learning (FL) \cite{mcmahan2017communication} offers a promising solution by enabling local training on the client side, where only model parameters or gradients are transmitted to a central server for aggregation, preserving data privacy and reducing communication costs.

Traditional synchronous federated learning (Sync-FL) faces inherent limitations, as the central server must await parameter updates from all selected clients before initiating model aggregation. This synchronous approach introduces the straggler effect \cite{vu2021straggler} due to heterogeneous computing capabilities, unreliable network connections, and dynamic environments. To address these challenges, asynchronous federated learning (Async-FL) \cite{xie2019asynchronous} has been introduced, where the central server aggregates as long as the model updates are received within a predefined time threshold in each round, enabling more flexible client participation.

However, Async-FL introduces significant challenges in managing stale gradients and model performance. In federated learning systems, the fundamental issue of client drift \cite{karimireddy2020scaffold} arises from non-independently and identically distributed (non-i.i.d.) data across clients \cite{hsu2019measuring}\cite{zhao2018federated}, where the global model can diverge from the optimum. This drift is intensified in Async-FL settings. Clients with superior computational capabilities and network conditions participate frequently in global model aggregation, whereas straggled clients contribute only sporadically. The resulting update imbalance amplifies the existing drift problem, creating a bias toward frequently participating clients. Furthermore, stale gradient information from straggled clients further impedes global model convergence. Dai et al. \cite{dai2018toward} analyzed the negative impact of stale gradients on convergence speed and suggests that the acceptable level of staleness in distributed training depends on model complexity. Zhou et al. \cite{zhou2022towards} proposed selecting stale gradients with a consistent descent direction to accelerate the overall training process. A semi-asynchronous federated aggregation (SAFA) approach \cite{wu2020safa} was proposed to aggregate and distribute the global model while allowing a certain latency. Yang et al. \cite{yang2020age} introduced age of information (AoI) to quantify the staleness of client updates and proposed a scheduling policy that accounted for both AoI and instantaneous channel quality. 
Ozfatura et al. \cite{ozfatura202212} leveraged the average AoI between clients as a regularization term to address exclusive client scheduling in channel-aware federated learning.

While the aforementioned studies assumed the availability of CSI and the successful reception of model parameters from clients on the central server, practical scenarios involving IoT devices operating in unlicensed frequency spectrum bands (FSB) face limitations in signal transmission capacity \cite{shariatmadari2015machine}. Furthermore, acquiring CSI via pilot transmission introduces additional signaling overhead, which increases as the number of clients grows \cite{raza2017low}. Several studies examined federated learning under unknown CSI. Amiri et al. \cite{amiri2021blind} considered the lack of CSI and designed a receiver beamforming scheme to compensate for it. In the context of federated edge learning, Razavikia et al. \cite{razavikia2024blind} employed q-ary quadrature amplitude modulation when the client does not have access to CSI. Tegin et al. \cite{tegin2023federated} studied federated learning for over-the-air computing and utilized multiple antennas to mitigate the effects of time-varying fading channels. To address the challenge of unknown channel distributions, some studies modeled multi-channel scheduling as a multi-armed bandit (MAB) problem, where users schedule channels according to a strategy and receive rewards based on transmission success. Notably, existing studies linked the cumulative reward to the freshness of the transmitted information. Bhandari et al. \cite{Bhandari2020} introduced the problem of minimizing the cumulative AoI in a single-source, stationary independent and identically distributed (i.i.d.) channel system, framing it as AoI bandits with AoI regret as the optimization metric. Qian et al. \cite{Qian2020} established policy-independent lower bounds on the average AoI for multiple users and channels. These studies assumed that the channel state follows a stationary but unknown distribution, or that the channel undergo smooth fading.
However, practical wireless environments exhibit inherently non-stationary scenario\cite{nguyen2023learning}, manifested through phenomena such as path fading, user mobility in telecommunication networks \cite{bian2021general}, and hostile jamming in extremely non-stationary communication scenarios \cite{mandal2022optimizing}. In non-stationary channels, the channel statistical distribution may change abruptly\cite{Pirayesh2022}. In federated learning systems, these non-stationary channels led to unpredictable client participation failures during global aggregation, which in turn increases the AoI of client updates. Mandal et al. \cite{mandal2022optimizing} studied the multi-users scheduling problem and proposed a greedy scheduling policy to minimize AoI of users in adversarial non-stationary environments. For piecewise-stationary channels, the Combinatorial Upper Confidence Bound (CUCB) algorithm \cite{chen2013combinatorial} was proposed to solve the combinatorial multi-armed bandit problem by balancing exploration and exploitation of unknown channels. Zhou et al. \cite{zhou2020near} proposed a change-point detector for detecting the moment when the channels undergo sudden changes.

Non-stationary channels can induce participation imbalance in aggregation, challenging federated learning fairness. This phenomenon was particularly prevalent in non-i.i.d. data scenarios, where the global model exhibited bias towards frequently participating clients with superior channel conditions and transmission success rates. Zhu et al. \cite{zhu2022online} proposed variance reduction to correct the inequality of participation rates of different clients. MAB-based client selection was used in \cite{ami2023client} to balance latency and generalization performance. Federated learning fairness covers both collaborative aspects and performance considerations\cite{mohri2019agnostic}\cite{kang2019incentive}. Collaborative fairness addresses contribution evaluation and reward allocation, considering the impact of non-i.i.d. data distribution on model convergence. Performance fairness targets model prediction bias reduction for enhanced cross-client generalization. Huang et al. \cite{huang2020efficiency} introduced a long-term fairness constraint for client selection to ensure that the average participation rate of each client did not fall below the expected guaranteed rate. Lyu et al. \cite{lyu2020collaborative} introduced a fairness evaluation framework that quantifies client reputation based on validation performance. The study in \cite{xu2020reputation} explored the discrepancy between local model updates and the global model, proposing this difference as an indicator of client reputation. To further refine fairness assessment, work in \cite{xu2021gradient} employed the cosine gradient Shapley value to approximate clients' marginal utility by analyzing the deviation between local and global updates. Jiang et al. \cite{jiang2023fair} focused on estimating the client's contribution using both gradient and data space to measure fairness in federated learning systems.

To the best of our knowledge, no research has explored the staleness and fairness problem of federated learning under the conditions where CSI is unavailable and the channels are non-stationary. We presents a MAB-based channel scheduling framework for federated learning in non-stationary wireless environments. The main contributions are summarized as follows:
\begin{itemize}
    \item For asynchronous federated learning in non-stationary scenarios, we first conduct theoretical analysis, utilizing AoI to quantify client staleness. Our work provides quantitative analysis on the impact of clients' AoI and successful participation rates and unifies these two factors to derive our optimization objective for channel scheduling. The channel scheduling problem under non-stationary channel conditions is analyzed within the MAB framework. 
    \item Our analysis encompasses two non-stationary channel scenarios: the piecewise-stationary case and the extremely non-stationary case. Building on this, we propose channel scheduling strategies based on GLR-CUCB and M-exp3 for these scenarios. Furthermore, we derive the achievable theoretical lower bounds on the AoI regret for both cases and establish the corresponding upper bounds, demonstrating their sublinear growth over time.
    \item To address the potential client update imbalance under non-stationary channel conditions, we propose an adaptive channel matching framework that considers priority coefficients based on marginal utility and fairness. Our algorithm dynamically adjusts channel matching in non-stationary environments, achieving more fair and efficient global aggregation. Experimental results demonstrate that our algorithm enhances communication efficiency and accelerates federated learning convergence.
\end{itemize}

The organization of this paper proceeds as follows: Section II introduces the asynchronous federated learning model under non-stationary channels. Section III presents the convergence analysis. Section IV establishes the AoI regret lower bounds for both extremely non-stationary and piecewise-stationary channels, and derives the upper bound under the proposed strategy. Section V details the adaptive channel matching algorithm. Section VI evaluates the experimental results. Section VII concludes our research.

\section{SYSTEM MODEL}
This section presents the asynchronous federated learning procedure and wireless channel model, focusing on two non-stationary channel types: extremely non-stationary and piecewise-stationary channels.
\subsection{Procedure of Federated Learning}
We consider a non-stationary wireless network consisting of \( M \) clients, indexed by the set \( \mathcal{M} = \{1, 2, \dots, M\} \), and a central server. Each client \( i \in \mathcal{M} \) holds a local dataset \( \mathcal{D}_i \), with \( |\mathcal{D}_i| \) representing the number of data samples. The clients collaborate with the central to complete privacy-preserving decentralized training. We denote the sample-wise loss function for client \(i\) as \( \mathcal{F}_i(\boldsymbol{w}_i, x) \), where \( \boldsymbol{w}_i \) represents the local model parameter of client \(i\), and \( x \in \mathcal{D}_i \) is a data sample from the dataset \( \mathcal{D}_i \). Then the local loss function for client \(i\) is given by:
\begin{equation}
    F_i(\boldsymbol{w}_i) = \frac{1}{|\mathcal{D}_i|} \sum_{x \in \mathcal{D}_i} \mathcal{F}_i(\boldsymbol{w}_i, x).
\end{equation}

Accordingly, the global loss function can be defined as:
\begin{equation}
    F(\boldsymbol{w}) = \frac{1}{M} \sum_{i=1}^M F_i(\boldsymbol{w}).
\end{equation}

The goal of federated learning is to determine the optimal global model parameters \( \boldsymbol{w}^* \) that minimize the global loss function:
\begin{equation}
    \boldsymbol{w}^* = \arg\min_{\boldsymbol{w}} F(\boldsymbol{w}).
\end{equation}

Due to the unpredictability and non-stationarity of CSI, only a subset of clients may successfully transmit their gradient updates to the central server in each round. We define \( S_t \) as the set of clients that successfully transmit their updates in round \( t \). Furthermore, we introduce AoI to quantify the degree to which clients are out-of-sync in the global aggregation process. Specifically, Let $a_i(t)$ denote the AoI of client $i$ in aggregation round $t$ and $h_i(t)$ denotes the most recent round in which client $i$ successfully participated in the global aggregation before round $t$. We have

\begin{equation}
    a_i(t) =  t - h_i(t).
\end{equation}

Then, the training process in round $t,\forall t\in{1,2,\cdots,T}$ can be summarized as follows:

\noindent$\bullet$ \textbf{(Step 1 Global Model Broadcast)} To mitigate computational resource wastage and optimize the utilization of local training outcomes, clients that failed to transmit their updates in round \( t-1 \) retain their gradient updates and do not participate in the new updates in round \( t \). Therefore, in round \( t \), the server only sends the latest global model parameters \( \boldsymbol{w}_t \) to the clients in \( S_{t-1} \). Given the sufficient transmission capacity of the central server, we assume that all transmissions over the downlink channel are always successful.

\noindent$\bullet$ \textbf{(Step 2 Local Model Update)} Each client \( i \in S_{t-1} \) initialize its local model  as \( \boldsymbol{w}_{i,t}^0 = \boldsymbol{w}_t \) and performs $E$ steps of stochastic gradient descent (SGD) for local model updating:
\begin{equation}
    \boldsymbol{w}_{i,t}^{e+1} =\boldsymbol{w}_{i,t}^{e} - \frac{\eta}{|\xi_{i,t}^e|} \sum_{x \in \xi_{i,t}^e} \nabla \mathcal{F}_i(\boldsymbol{w}_{i,t}^{e}, x),
\end{equation}
where $e\in\{0,1,\cdots,E-1\}$, \( \eta \) is the learning rate and \( \xi_{i,e} \subseteq \mathcal{D}_i \) is a mini-batch of data samples selected during iteration \( e \).

\noindent$\bullet$ \textbf{(Step 3 Local Update Uploading)}
After each client \( i\in S_{t-1} \) completes \( E \) steps local iterations, all clients in $\mathcal{M}$ update its latest cumulative update as
\begin{equation}
\tilde{\boldsymbol{G}}_{i,t} =
\begin{cases}
\frac{1}{\eta} \left( \boldsymbol{w}_{i,t}^0 - \boldsymbol{w}_{i,t}^E \right). & \text{if } i \in S_{t-1} \\
\tilde{\boldsymbol{G}}_{i,t-1}. & \text{otherwise}
\end{cases}
\end{equation}
Clients that failed to participate in the previous round of global aggregation will pause local updates until they successfully transmit their latest parameters to the central server. The central server assigns each client a channel to transmit its model update $\tilde{\boldsymbol{G}}_{i,t}$ to the server.

\noindent$\bullet$ \textbf{(Step 4 Global Model Update)}
Only the model updates from clients with favorable channel conditions are successfully transmitted to the server. The global model can thus be updated as follows:
\begin{equation}
    \boldsymbol{w}_{t+1} = \boldsymbol{w}_{t}-\frac{1}{|S_t|} \sum_{i\in S_t}\zeta_{i}^{t}\tilde {\boldsymbol{G}}_{n,t},
\end{equation}
where $\zeta_{i}^{t}$ denotes the aggregation weight of client $i$ in round $t$. We rewrite Equation (4) and consequently the AoI of each client $i,\forall i\in\mathcal{M}$ is updated as
\begin{equation}
a_i(t) =
\begin{cases}
1. & \text{if } i \in S_t \\
a_i(t-1) + 1. & \text{otherwise}
\end{cases}
\end{equation}
We assumpt that $a_i(0)= 1$ for all clients $i \in \{1 , 2, \cdots, M \} $ in round $ =0$. The above steps are repeated until round \(T\) is reached.

\subsection{Non-stationary Wireless network}
In this wireless network, we equally divide the spectrum into \( N \) orthogonal sub-channels (\( N \geq M \)), indexed by the set \( \mathcal{N} = \{1, 2, \dots, N\} \).  Each client periodically attempts to transmit the time-sensitive local updates to the central server over one of the \( N \) channels.

The central server schedules distinct sub-channels to clients based on a scheduling policy, preventing collisions within each round. Sub-channels are modeled as Bernoulli channels with state Good (1) or Bad (0) at any round $t$. Clients cannot obtain real-time CSI or prior knowledge of the channels' statistical properties.

In the stationary setting, channel states are independent between channels over $T$ rounds. Let $\mu_k$ denote the mean state of channel $k \in \{1, 2, \cdots, N\}$, which is unknown to client $i \in \{1, 2, \cdots, M\}$ and constant throughout $T$. Without loss of generality, we assume $\mu_1 > \mu_2 > \cdots > \mu_N$, where $\{1, \cdots, M\}$ are the $M$-best channels and $\{M+1, \cdots, N\}$ are the $N-M$-worst channels.

In the non-stationary setting, channel states may fluctuate due to factors like malicious attacks or collisions \cite{mandal2022optimizing}. In the extremely non-stationary scenario, an adversary pre-determines the state sequence of each channel (Good or Bad) without relying on any statistical assumptions\cite{nguyen2017impact}. For piecewise-stationary channels, we assume the central server has some prior knowledge, where the mean channel state \( \mu_k \) for each \( k = 1, 2, \dots, N \) remains constant over certain intervals and changes only at unknown rounds.

\section{Problem Formulation And Convergence Analysis}
\subsection{Problem Formulation}

The goal of this study is to design an effective online channel scheduling strategy for non-stationary environments with unknown CSI. This strategy enables clients to identify reliable channels, ensuring successful transmission of parameters to the server in federated learning, thereby accelerating convergence and enhancing the generalization performance of the global model. Based on this, the problem is formulated as follows:
\begin{subequations}
\renewcommand{\theequation}{9\alph{equation}}
\begin{align}
\mathcal{P}: &\min_{\{\boldsymbol{\beta}_{t}\}_{t=1}^T}F(\boldsymbol{w}_T) \tag{9} \\
\text{s.t. }& \sum\nolimits_{k=1}^N\beta_{i,k}^t=1, \quad\forall i \in \mathcal{M},\forall t, \tag{9a} \\
&\sum\nolimits_{i=1}^M\beta_{i,k}^t\leq 1, \quad\forall k \in \mathcal{N},\forall t, \tag{9b} 
\end{align}
\end{subequations}
where $\beta_{i,k}=1$ if sub-channel $k$ is allocated to client $i$, and $\beta_{i,k}=0$ otherwise. $(9a)$ means each client is assigned a channel, and $(9b)$ constraints that each channel is assigned to at most one client.

Problem $\mathcal{P}$ remains challenging to solve because the factors influencing the performance of federated learning are still unclear. The CSI is unknown and the channel state exhibits non-stationary distribution. To address the aforementioned challenges, we begin by performing a convergence analysis of federated learning in non-stationary transmission scenarios, emphasizing the key factors that influence model performance. Based on this analysis, we can then design our channel scheduling strategy. 
\subsection{Convergence Analysis}
Before conducting the convergence analysis, we first present the necessary assumptions.

\noindent$\textbf{\textit{Assumption 1.($L$-Smooth)}}$ For all clients $i \in \{1 , 2, \cdots, M \}$, the local loss function is L-soooth, i.e., ${F_i}(a) - {F_i}(b) \le \left \langle  \nabla F_i (b),(a - b)\right \rangle  + \frac{L}{2}\left\| {a - b} \right\|^2$.

\noindent$\textbf{\textit{Assumption 2.(Bounded local gradient)}}$ For each client \( i \in \{1, 2, \dots, M\} \), communication round \( t \in \{1, 2, \dots, T\} \), and local training epoch \( e \in \{1, 2, \dots, E\} \), the expected squared norm of the stochastic local gradients is uniformly bounded, i.e., $\mathbb{E}{\left\| {\nabla {F_i}(\boldsymbol{w} _{i,t}^e| \xi_{i,t}^e)} \right\|^2} \le {G^2}$.

\noindent$\textbf{\textit{Assumption 3.(Bounded dissimilarity)}}$
Given the global loss function $F({\boldsymbol{w} _t})$ and local loss function $F_i({\boldsymbol{w} _t})$, for all clients $i \in \{1 , 2, \cdots, M \}$, we have $\mathbb{E}{\left\| {\nabla F({\boldsymbol{w} _t}) - \nabla {F_i}({\boldsymbol{w} _t})} \right\|^2} \le {\delta ^2}$. 

Based on the aforementioned assumptions and Equation (4), we employ AoI analysis to examine client staleness. We first derive Theorem 1, which analyzes the impact of both the AoI of the clients and the number of participating clients in each round of aggregation on model convergence. These two factors are subsequently integrated, followed by the introduction of the problem transformation.
\begin{theorem}
Given $\eta<\frac{1}{9L}$, after \(T\) rounds of training, the difference between the loss of the global model and the optimal loss can be bounded as
\begin{equation}
\begin{aligned}
\mathbb{E}\left[F(\boldsymbol{w}_T)-F(\boldsymbol{w}^*)\right]&\leq\Omega^T\mathbb{E}\left[F(\boldsymbol{w}_0)-F(\boldsymbol{w}^*)\right]\\
&+\alpha_1\sum_{t=1}^{T-1}\Omega^{T-1-t}\left ( 1-\frac{|S_t|}{M}\right  )^2G^2\\
&+\alpha_2\sum_{t=1}^{T-1}\Omega^{T-1-t}(\frac{1}{M} \sum_{i=1}^Ma_i(t))^2\\
&+\alpha_3\frac{1-\Omega^T}{1-\Omega}, 
\end{aligned}
\end{equation}
\end{theorem}
where $\alpha_1=6L\eta^2\lambda^2+2\eta\lambda$, $\alpha_2=9KL^3\eta ^4\lambda ^4(\sigma ^2+G^2)$, $\alpha_3=\frac{3L \eta^2\lambda^2 \sigma^2}{2}+\frac{3L^3\eta^2G^2\lambda^3(\lambda-1)(2\lambda -1)}{4}$ and $\Omega=1-\eta \lambda L+9\eta^2L^2\lambda $.

\noindent\textit{Proof:} See the proof in Appendix A.

\noindent\textbf{Remark 1:} Theorem 1 states that the loss of the global model is primarily influenced by two factors. The second term $\alpha _1\left ( 1-\frac{|S_t|}{M}\right  )^2G^2$ quantifies the impact of the number of clients successfully participating in training in each round. Specifically, a larger \( |S_t| \) reduces the second term, thereby decreasing the convergence error and accelerating convergence. When all clients participate successfully, i.e. $|S_t|= M$, this term becomes zero. The third term $\alpha_2(\frac{1}{M} \sum_{i=1}^M a_i(t))^2$ reflects the influence of the clients' AoI, which represents the effect of outdated gradient information. Reducing $\frac{1}{M} \sum_{i=1}^Ma_i(t)$, i.e. the average AoI of all clients will also decrease the upter bound and accelerate convergence.

\noindent\textbf{Remark 2 :} To ensure the model converges, we require \( \Omega < 1 \), which implies \( \eta < 9L \). As \( T \to \infty \), the first term \( \Omega^T \mathbb{E}[F(\boldsymbol{w}_0) - F(\boldsymbol{w}^*)] \) tends to 0. However, the other three terms remain, representing the gap between the global model and the optimal model. To reduce this gap, we can improve model performance by decreasing the second and third terms, which means increasing \( |S_t| \) and decreasing the average AoI of clients, i.e. \( \frac{1}{M} \sum_{i=1}^M a_i(t) \).
Next, we will demonstrate in Lemma 1 that the average AoI of clients is inversely related to \( |S_t| \), enabling a unified analysis of these two factors.
\begin{lemma}
Assume the network contains $M$ clients, and $|S_t|$ clients successfully participate in the global model training each round. Then, the clients' average AoI satisfies the following condition:
\begin{equation}
\mathbb{E}[AoI]=\frac{M^2}{|S_t|}.
\end{equation}
\end{lemma}
\noindent\textit{proof}: With uniform client selection probability, we analyze client $i$'s AoI. The probability of selecting any client is $p = \frac{S_t}{M}$ with $S_t$ clients selected per round. For client $i$, the probability of $AoI = j$ is:
\begin{equation}
P(AoI_{t,i}=j)=p(1-p)^{j-1}.
\end{equation}
This means that client $i$ has passed $j$ rounds since its last successful transmission. Therefore, the expected AoI of client $i$ in round $t$ is
\begin{equation}
\begin{aligned}
&\mathbb{E}[AoI_{t,i}] \\
&= \sum_{j  = 1}^{\infty } jP(AoI_{t,i}  = j)\\
&=\lim_{t \to \infty } \frac{p}{1-p}\left [ (1-p)+2(1-p)^2+\cdots +j(1-p)^j  \right ]\\
&=\lim_{t \to \infty}\frac{p}{1-p} \left [\frac{(1-p)[1-(1-p)^j]}{p^2}-j(1-p)^{j+1}\right ]\\
&=\frac{1}{p}=\frac{M}{|S_t|}.
\end{aligned}
\end{equation}
Therefore, the overall network AoI is $\mathbb{E}[AoI_{t}]=\frac{M^2}{|S_t|}$. This completes the proof.

Lemma 1 shows client average AoI negatively correlates with transmission success set size $|S_t|$. Minimizing client AoI is equivalent to maximizing successful transmissions $|S_t|$ per round, both contributing to model convergence acceleration. Based on these theoretical results, we transform optimization problem $\mathcal{P}$ in the following subsection.

\subsection{Problem Transformation}

As previously mentioned, the model's performance and convergence speed are influenced by  AoI of the clients. Therefore, selecting good channels for gradient transmission is crucial. However, due to the time-varying nature of the channel's statistical properties and the unknown CSI, existing methods can not be directly applied to solve this issue. To overcome these challenges, we leverage the MAB-based framework to design an online channel scheduling strategy that ensures reliable gradient transmission for clients, thus minimizing the cumulative AoI of all clients during \( T \) rounds of training. Suppose there exists an oracle strategy that has access to the instantaneous CSI, we reformulate the problem as to design a MAB-based channel scheduling strategy to minimize the AoI regret $R_{\pi}(T)$, which is defined as follows:
\begin{equation}
    \min R_{\pi}(T)=\sum_{i = 1}^M \sum_{t = 1}^T \mathbb{E}[a_i^{\pi}(t) - a_i^*(t)],
\end{equation}
where \( a_i^{\pi}(t) \) and \( a_i^*(t) \) denote the AoI of client \( i \) at the \( t \)-th round under the channel scheduling strategy \( \pi \) and the oracle strategy, respectively. \( R_{\pi}(T) \) represents the total AoI regret of strategy \( \pi \) over rounds 1 to \( T \). The expectation $\mathbb{E}[\cdot]$ is taken with respect to both the randomization over channel configuration and the policy. Without loss of generality, we assume that $a_i^{\pi}(0)= a_i^*(0) = 1$ for all clients $i \in \{1 , 2, \cdots, M \} $ in round $t =0$. 

Given the dynamic fluctuations of channel statistics across rounds and the probabilistic success of transmissions, quantifying the number of successful client updates per round presents significant challenges. Our scheduling policy for non-stationary environments aims to minimize the overall AoI, while also incorporating fairness considerations to ensure effective generalization performance. 

In the following, we will present the channel scheduling strategy for non-stationary environments, alongside a fairness-aware channel matching approach, both of which collaboratively constitute the overall framework for asynchronous federated learning in such scenarios.

\section{CHANNEL SCHEDULING IN NON-STATIONARY ENVIRONMENTS}
In this section, we model two types of non-stationary channel environments: extremely non-stationary channels and piecewise-stationary channels. For each environment, we design channel scheduling strategies to minimize AoI regret. Additionally, we conduct theoretical analysis of the performance of the proposed algorithms.

\subsection{Channel scheduling in extremely non-stationary environment}
In the case of extremely non-stationary channels, we model the channel scheduling problem in federated learning as a Multi-Player Multi-Armed Bandit (MP-MAB) problem \cite{Uchiya2010}. Each client is viewed as a player, and each non-stationary channel is viewed as an arm, while the central server acts as the coordinator responsible for selecting and allocating sub-channels. 
Before introducing the channel scheduling strategy, we first theoretically derive the lower bound of AoI regret that can be achieved in such an environment. We first present the following lemma.
\begin{lemma}
Let $s_i(t)$ denote the index of the channel allocated to client $i$ in round $t$, and each channel $s_i(t)$ subject to the Bernoulli distribution $\mathcal{B}_{\mu_{s_i(t)}}$, where $\mu_{s_i(t)}$ denotes the mean value. $\mathbb{E}[a_i(t)]$ denotes the expected AoI of client $i$ in round $t$, then there are
\begin{equation}
    \mathbb{E}[a_i(t)] = \sum\limits_{\tau  = 0}^\infty  {\prod\limits_{k = 0}^\tau  {(1 - {\mu _{s_i(t - k)}})} } .
\end{equation}
\end{lemma}
\noindent\textit{Proof}:
From the definition of the AoI, 
\begin{equation}
    \mathbb{P}(a_i(t) > \tau ) = \prod\limits_{k = 0}^\tau  {(1 - {\mu _{s_i(t - k)}})}.
\end{equation}
Thus we have
\begin{equation}
    \mathbb{P}(a_i(t) = \tau ) = \mathbb{P}(a_i(t)) > \tau  - 1) - \mathbb{P}(a_i(t) > \tau ),
\end{equation}
\begin{equation}
\begin{split}
\mathbb{E}[a_i(t)] &= \sum\limits_{\tau  = 0}^\infty  {\tau \mathbb{P}(a_i(t) = \tau )} \\
 &= \sum\limits_{\tau  = 0}^\infty  {\tau (\mathbb{P}(a_i(t) > \tau  - 1)}  - \mathbb{P}(a_i(t) > \tau )) \\
 &= \mathbb{P}(a_i(t) > 0) - \mathbb{P}(a_i(t) > 1) \\ &+2(\mathbb{P}(a_i(t) > 1) - \mathbb{P}(a_i(t) > 2)) +  \cdots \\
 &= \sum\limits_{\tau  = 0}^\infty  {\mathbb{P}(a_i(t) > \tau )} \\
 &= \sum\limits_{\tau  = 0}^\infty  {\prod\limits_{k = 0}^\tau  {(1 - {\mu _{s_i(t - k)}})} }.
\end{split}
\end{equation}

This completes the proof.

Based on Lemma 2, we now give the lower bound of AoI regret that can be achieved by any channel scheduling strategy under extremely non-stationary channel environments:
\begin{theorem}
For federated learning in an extremely non-stationary channel environment, when there are \( M \) participating clients, \( N \geq M \) channels, and \( T \) aggregation rounds, there exists a channel state distribution such that the AoI regret of any strategy is lower bounded by \( \Omega \left( \frac{(N-M)^2}{N^2} \sqrt{NT} \right) \).
\end{theorem}
\noindent\textit{Proof}: See the proof in Appendix B.

Theorem 2 reflects the impact of the number of clients \( M \) and the number of channels \( N \) on the performance of federated learning. When the number of clients \( M \) is fixed, a larger number of available channels \( N \) increases the size of the super arm, reducing the probability of exploring the optimal super arm in each round. This results in a lower probability of client gradient updates successfully transmitted to the central server, thereby increasing the client's AoI and weakening the model performance.

To enable the central server to identify optimal channels, we extend the Exp3.S algorithm \cite{Auer2002} and propose the M-Exp3 algorithm tailored for federated learning in extremely non-stationary channels. Specifically, the M-Exp3 algorithm treats the \( M \) clients as a group of super players and the combinations of \( M \) channels as super arms. In each round, the algorithm probabilistically schedules different channel combinations and adjusts the scheduling probabilities for the subsequent round based on the gradient reception results from the server. The detailed steps of the algorithm are outlined in Algorithm 1.
We now analyze the performance of the M-Exp3 algorithm theoretically, as stated in Theorem 3.
\begin{algorithm}[H]
	\caption{Multi-player of Exp3 (M-Exp3)} 
	\label{algorithm} 
	\begin{algorithmic}[1]
   \REQUIRE $T, C = |C(N,M)|$ and $\gamma \in (0, 1]$
      \STATE Initialization:$w_I(0) = 1$ for all $I \in C(N,M)$ 
		\WHILE{$t \leq T$} 
       \STATE  Set
\begin{equation}
p_I(t) = (1-\gamma) \frac{w_I(t)}{\sum_{J \in C(N,M)} w_J(t)}  + \frac{\gamma}{C}, \ \ \  I \in C(N,M)
\nonumber
\end{equation}
\STATE Draw super-arm $I^t$ according to the probabilities $[p_J(t)]_{J\in C(N,M)}$.
\STATE Receive super-rewards $X_t(I^t) = \sum_{i \in I^t} X_t(i)$
\STATE For $J \in C(N,M)$ set
\begin{equation}
\hat{X}_t(J)=\left\{
\begin{array}{cl}
X_t(J)/ p_{J}(t)   &\textup{if} \ \  J = I^t\\
0   & \textup{otherwise} \\
\end{array} \right.
\nonumber
\end{equation}
\begin{equation}
w_{J}(t+1) = w_J(t) \exp \left(\frac{\gamma \hat{X}_t(J)}{C} \right)
\nonumber
\end{equation}
		\ENDWHILE 
	\end{algorithmic} 
\end{algorithm}
\begin{theorem}
In federated learning scenarios with extremely non-stationary environments, where the M-Exp3 algorithm is employed to select the channels between clients and the server, the cumulative AoI regret over \( T \) rounds is bounded by \( \mathcal{O}(M|C(N,M)|^2 \sqrt{T|C(N,M)| \log |C(N,M)|}) \), where $C(N,M)$ denotes all combinations of selecting $M$ channels from $N$ available channels.
\end{theorem}
\noindent\textit{Proof}: Let $G_{\max}^1(T), \dots, G_{\max}^M(T)$ denote the number of successful transmissions (Good state of the channel selected) of each client by the oracle policy with $G_{\max}(T) = \sum_{m =1}^M G_{\max}^m(T)$, and let $G_{\textup{M-Exp3}}^1(T), \dots, G_{\textup{M-Exp3}}^M(T)$ denote the corresponding number for M-Exp3 for the $M$ players in $T$ rounds of aggregation with $G_{\textup{M-Exp3}}(T) = \sum_{m =1}^M G_{\textup{M-Exp3}}^m(T)$.
From \cite[Corollary 3.2]{Auer2002}, 
\begin{equation}
\begin{aligned}
&G_{\max} -  \mathbb{E}[G_{\textup{M-Exp3}}] \\
&\hspace{2cm}\leq2M\sqrt{e - 1}\sqrt{T |C(N,M)| \log |C(N,M)|}.
\end{aligned}
\end{equation}
For player $i$, we have $\mu^*_i =  \frac{G^i_{\max}}{T} $ for oracle policy and  $\mu_i =  \frac{\mathbb{E}[G^i_{\textup{M-Exp3}}]}{T} $ for M-Exp3, then from (14),
\begin{align}
&R_{\pi}(T)  = \sum_{t = 1}^T \sum_{i=1}^M \mathbb{E}\left[ \frac{T}{G_{\textup{M-Exp3}}^i} -  \frac{T}{G_{\max}^i} \right] \notag\\
&= T^2 \left(\frac{G_{\max}^1 - G_{\textup{M-Exp3}}^1}{G_{\max}^1G_{\textup{M-Exp3}}^1}  + \cdots +  \frac{G_{\max}^M - G_{\textup{M-Exp3}}^M}{G_{\max}^MG_{\textup{M-Exp3}}^M}  \right) \notag\\
& \leq T^2 \left( \frac{G_{\max} - G_{\textup{M-Exp3}}}{G_{\max}^j G_{\textup{M-Exp3}}^j}\right)\notag\\
& \leq  \frac{T^2 (G_{\max} - G_{\textup{M-Exp3}})}{G_{\max}^j (G_{\max}^j - 2\sqrt{e - 1}\sqrt{T |C(M,N)| \log |C(M,N)|})},
\end{align}
where $G_{\max}^j G_{\textup{M-Exp3}}^j = \min_{i\in [M]} G_{\max}^i G_{\textup{M-Exp3}}^i$. 
If  $G_{\max}^j \geq \frac{T}{|C(N,M)|}$ and  $T > 16(e-1)|C(N,M)|^3 \log |C(N,M)|$, then
\begin{equation}
\begin{aligned}
&R_{\pi}(T)\leq  4 M \sqrt{e - 1} |C(N,M)|^2\\
&\hspace{3cm}   \times  \sqrt{T|C(N,M)| \log |C(N,M)|}.
\end{aligned}
\end{equation}
\subsection{Channel scheduling in piecewise-stationary environment}
Piecewise-stationary channel is a weakened form of extremely non-stationary channel, where the channel mean $\mu_i,\forall i\in\mathcal{N}$ of each channel remains constant over several consecutive rounds but may experience abrupt changes at certain rounds. When $\mu$ undergoes a sudden change in a round compared to the previous round, we refer to this round as a breakpoint. Therefore, the channel scheduling problem under piecewise-stationary channels can be modeled as a multi-player piecewise combinatorial multi-arm problem\cite{chen2013combinatorial}. We analyze the minimum achievable AoI regret under such channel conditions.
\begin{theorem}
With the piecewise-stationary channel scheduling problem modeled as multi-player combinatorial semi-bandits, for any number of users $M$ , for any number of channels $ N \geq M$ and for any aggregation rounds $T$, there exists a distribution over the assignment of channel states such that the AoI regret of any policy is $\Omega (\sqrt{\frac{MT}{N}})$.
\end{theorem}
\textit{Proof}: See the proof in Appendix C.

Since the evolution of AoI in each stationary interval under a piecewise-stationary channel is similar to that under a stationary channel, we first derive Lemma 3 and Lemma 4, which pertain to multi-channel scheduling under stationary channels. For analytical convenience, we assume an idealized scenario where each client takes turns using the optimal channel during \(T\) rounds of training.
\begin{lemma}
In the stationary channel, let $k_i(t)$ be the index of the channel scheduled to client $i$ in round $t$, $k_i^*(t)$ be the index of the optimal channel scheduled to client $i$ in round $t$ by oracle policy, and $a_i(t)$ be AoI of client $i$ in round $t$, such that ${\mu _{\min }} = \mathop {\min }\limits_{k \in [N]} {\mu _k}$, $c = \frac{{ - 1}}{{\log \left( {\prod\limits_{k = 1}^M {(1 - {\mu _k})} } \right)}}$ and $c' = Mc$, we have,
\begin{multline}
\sum\limits_{t = 1}^T  \mathbb{E}[{a_i}(t)] \le Q + \frac{{c'\log T + 1}}{{{\mu _{\min }}}} + \frac{{c'\log T}}{{{\mu _{\min }}}}\mathbb{E}[ {\sum\limits_{t = 1}^T {{\mathbb{I}_{{k_i}(t) \ne k_i^*(t)}}} } ],
\end{multline}
where $Q = \left( {\sum\limits_{\tau  = 0}^{c'\log T} {\prod\limits_{m = 0}^\tau  {(1 - {\mu _{k_i^*(t - m)}})} } } \right)(T - c'\log T)$.
\end{lemma}
\textit{Proof:} When $t > Mc\log T$, let $E_t$ represent the event: for $t - Mc\log T+ 1\le\tau \le t$, ${k_i}(\tau) = k_i^*(\tau)$, i.e. ${E_t}=\bigcap\nolimits_{\tau = t - Mc\log T+1}^t {{k_i}(\tau) = k_i^*(\tau)}$. Let $E_c^t$ denote the event: $E_t^c = \bigcup\nolimits_{\tau = t-Mc\log T + 1}^t {{k_i}(\tau ) \ne k_i^*(\tau )}$, then
\begin{equation}
\begin{split}
\mathbb{E}[{a_i}(t)|{E_t}] &= \mathbb{E} \left[ \sum\limits_{\tau  = 0}^\infty  {\prod\limits_{m = 0}^\tau  {(1 - {\mu _{{k_i}(t - m)}})} } |{E_t}] \right]\\
&\le \sum\limits_{\tau  = 0}^{Mc\log T}{\prod\limits_{m = 0}^\tau  {(1 - {\mu _{k_i^*(t - m)}})}} \\
&+\sum\limits_{\tau  = Mc\log T + 1}^\infty  {{\left( {\prod\limits_{k = 1}^M {(1 - {\mu _k})} } \right)}^{c\log T}} \\
&\prod\limits_{m = Mc\log T + 1}^\tau(1 - {\mu _{\min }}) \\
&\le \sum\limits_{\tau  = 0}^{Mc\log T} {\prod\limits_{m = 0}^\tau  {(1 - {\mu _{k_i^*(t - m)}})}  + } \frac{1}{{{\mu _{\min }}T}},\\
\end{split}
\end{equation}
where $c = \frac{{ - 1}}{{\log \left( {\prod\limits_{k = 1}^M {(1 - {\mu _k})} } \right)}}$. And we have
\begin{equation}
\begin{split}
    \mathbb{E}[{a_i}(t)|E_t^c] &= \mathbb{E}\left[\sum\limits_{\tau  = 0}^\infty  {\prod\limits_{m = 0}^\tau  {(1 - {\mu _{{k_i}(t - m)}})} } |E_t^c \right] \\
    &\le \frac{1}{{{\mu _{\min }}}}.
\end{split}
\end{equation}
Let $c' = Mc$, we have
\begin{multline}
\begin{split}
\sum\limits_{t = 1}^T  \mathbb{E}[{a_i}(t)] &= \sum\limits_{t = 1}^{c'\log T}  \mathbb{E}[{a_i}(t)] 
+ \sum\limits_{t = c'\log T + 1}^T  \mathbb{E}[{a_i}(t)] \\
&\le \frac{c'\log T}{{\mu _{\min }}} + \sum\limits_{t = c'\log T + 1}^T \Bigl(\mathbb{P}({E_t})\mathbb{E}[{a_i}(t)|{E_t}] \\
&\quad+ \mathbb{P}(E_t^c)\mathbb{E}[{a_i}(t)|E_t^c] \Bigr) \\
&\le \frac{c'\log T}{{\mu _{\min }}} + \Biggl( \sum\limits_{\tau  = 0}^{c'\log T} \prod\limits_{m = 0}^\tau  (1 - {\mu _{k_i^*(t - m)}}) \\
&\quad+ \frac{1}{{\mu _{\min }T}} \Biggr)(T - c'\log T) \\
&\quad+ \frac{c'\log T}{{\mu _{\min }}} \mathbb{E} \left[ \sum\limits_{t = 1}^T {\mathbb{I}_{{k_i}(t) \ne k_i^*(t)}} \right].
\end{split}
\end{multline}
This completes the proof.


In federated learning under a piecewise-stationary channel environment, channels that were previously in good condition may suddenly degrade, leading to failed gradient update transmissions from clients to the server. To address this issue, we use the Generalized Likelihood Ratio (GLR) as the breakpoint detector and the Combinatorial Upper Confidence Bound (CUCB) as the scheduling algorithm\cite{zhou2020near}. When no breakpoint is detected, CUCB is used for channel scheduling within each stationary interval. During stationary channel conditions, the UCB value from the previous round is incorporated into the decision for each scheduling round. When a breakpoint is detected, the CUCB algorithm is restarted. The GLR-CUCB algorithm is presented in Algorithm 2 and we have the following lemma.

\begin{lemma}
Within the $T$ rounds of aggregation, the piecewise-stationary channel can be partitioned into $C_T$ intervals, with each interval treated as a stationary channel. The expected total number of occurrences of non-optimal scheduling events under the GLR-CUCB strategy is bounded by ${\cal O}(\sqrt {C_T NT \log T})$ when $C_T$ is known, and by ${\cal O}(C_T \sqrt {NT \log T})$ when $C_T$ is unknown.
\end{lemma}

\noindent\textit{Proof:}
Let $k(t)$ denote the set of $M$ arms selected by the GLR-CUCB policy over time round $t$, and $k^*(t)$ denote the set of $M$ arms optimally selected by the optimal policy over time round $t$, there are
\begin{equation}
    \begin{split}
        \sum\limits_{t = 1}^T {{\mathbb{I}_{{\textbf{k}}(t) \ne \textbf{k}^*(t)}}}  &= \sum\limits_{t = 1}^T {\frac{{{\Delta _{\min }}{\mathbb{I}_{\textbf{k}(t) \ne {\textbf{k}^*}(t)}}}}{{{\Delta _{\min }}}}} \\ 
        &\le \sum\limits_{t = 1}^T {\frac{{\Delta (t){_{\textbf{k}(t) \ne {\textbf{k}^*}(t)}}}}{{{\Delta _{\min }}}}}  \\
        &= \frac{{R(T)}}{{{\Delta _{\min }}}},
    \end{split}
\end{equation}
where $\Delta (t) = \sum\nolimits_{k \in {\textbf{k}^*}(t)} {{\mu _k}}  - \sum\nolimits_{k \in \textbf{k}(t)} {{\mu _k}} $ and ${\Delta _{\min }} = {\min _{t \in [T]}}\Delta (t)$. $R(T)$ denote the accumulated regret of GLR-CUCB. With the \cite[Corollary 4.3]{zhou2020near}, we arrive at Lemma 4.

Based on Lemma 3 and Lemma 4, we can analyze the performance of the GLR-CUCB channel scheduling algorithm, as stated in Theorem 5.

\begin{theorem}
For federated learning in a piecewise-stationary channel environment, when $C_T$ is known, the AoI regret value of GRL-CUCB is $\mathcal{O}(M\sqrt{C_T N T \log^3 T})$, and when $C_T$ is unknown, the AoI regret value of GRL-CUCB is $\mathcal{O}(MC_T\sqrt{ N T \log^3 T})$.
\end{theorem}
\textit{Proof}: Let $i^*(t)$ be the index of the the $i$-th largest channel of player $i$ in round $t$. 
Let $\tau_1, \tau_2, \cdots, \tau_{C_T - 1}$ denote the breakpoints, then $t \in \{1,2,\cdots, \tau_1\},\{\tau_1+1,\tau_1+2,\cdots, \tau_2\},\cdots, \{\tau_{C_T-1}+1,\tau_{C_T-1}+2,\cdots, T\}$ can be divided to $C_T$ intervals, where $\tau_0 = 0$, $\tau_{C_T} =T$. Let $\mu_{\min}$ denote the smallest means of all intervals amonge all arms. Because each interval is stationary, Lemma 3 can be used directly for $j$-th interval. From Lemma 3,
\begin{equation}
\begin{split}
\sum_{t=\tau_j + 1}^{\tau_{j + 1}} \mathbb{E}[a_i(t)]  &\leq \frac{\tau_{j + 1} - \tau_{j}}{\mu_{i^*(t)}} + \frac{c \log (\tau_{j + 1} - \tau_{j}) + 1}{\mu_{\min}}  + \\
& \frac{c \log (\tau_{j + 1} - \tau_{j})}{\mu_{\min}} \mathbb{E} \left[ \sum_{t=\tau_j + 1}^{\tau_{j + 1}} \mathbb{I}_{k_i(t) \neq i^*(t)} \right].
\end{split}
\end{equation}
The AoI regret of all players  is,
\begin{equation}
\begin{split}
&R_\pi(T)= \sum_{j = 0}^{C_T - 1} \sum_{t=\tau_j + 1}^{\tau_{j + 1}} \sum_{i = 1}^{M} \mathbb{E}[a_i(t) - a^*_i(t)] \\
 &\leq  \sum_{j = 0}^{C_T -1} \sum_{i=1}^M \frac{c \log T + 1}{\mu_{\min}} + \\
& \sum_{j = 0}^{C_T -1} \sum_{i=1}^M  \frac{c \log T}{\mu_{\min}} \mathbb{E} \left[ \sum_{t=\tau_j + 1}^{\tau_{j + 1}} \mathbb{I}_{k_i(t) \neq i^*(t)} \right] \\
&  \leq  \frac{C_T M  (c \log T + 1)}{\mu_{\min}} +  \frac{c \log T}{\mu_{\min}} \mathbb{E} \left[ \sum_{i=1}^M  \sum_{t=1}^{T} \mathbb{I}_{k_i(t) \neq i^*(t)} \right] \\
&  \leq  \frac{C_T M  (c \log T + 1)}{\mu_{\min}} +  \frac{c \log T}{\mu_{\min}} \mathbb{E} \left[ M  \sum_{t=1}^{T} \mathbb{I}_{\textbf{k}(t) \neq \textbf{k}^*(t)} \right]. \\
\end{split}
\end{equation}
From Lemma 4, we arrive at Theorem 5.
This completes the proof.

\begin{algorithm} [htbp]
	\caption{ Combinatorial Upper Confidence Bound Algorithm with Generalized Likelihood Ratio(GLR-CUCB)} 
	\label{alg1} 
	\begin{algorithmic}[1]
   \REQUIRE $T, M, \alpha, \delta $ and $N$
      \STATE Initialization: $\tilde{\mu}_i = 0, D_i = 0, a_j = 0  \  \forall \  i \in [N],  j \in [M] $, $\tau = 0$
		\WHILE{$t \leq T$} 
		\IF{$ \alpha \geq 0$ and $ i \gets (t - \tau)$ mod $ \lfloor {\frac{N}{\alpha} } \rfloor \in [N]$}
        \STATE  Randomly choose $S_{t+1}$ with  $ i  \in  S_{t+1}$ 
      \ELSE
        \STATE For each arm $i$, update 
        \STATE 
\begin{equation}
 \hat{\mu}_i(t) = \tilde{\mu}_i(t-1) + \sqrt{\frac{3 \log (t - \tau)}{ 2 D_i(t-1) }}
\end{equation}
        \STATE Choose $S_{t+1}$ with  the $M$ largest $\hat{\mu}_i(t)$
      \ENDIF 
      \STATE For each player $j$, choose the $ ((j + t)$  mod $ M )$-th largest in $S_{t+1}$
  
     \STATE  Get reward $X_i(t),  \forall \  i \in S_{t+1} $
     \STATE Update  $a_j(t), \forall \  j \in [M] $
     \STATE $\tilde{\mu}_i(t) \gets \frac{\tilde{\mu}_i(t-1) D_i(t-1) + X_i(t)}{ D_i(t) + 1},  \forall \  i \in S_{t+1} $
     \STATE $D_i(t) \gets D_i(t-1) + 1,  \forall \  i \in S_{t+1} $
      \STATE  \textit{Run GLR Change-Detection Detector:} 
      \FOR{ $\forall \  i \in S_{t+1}$}
      \STATE  $Z_{i, D_i(t)} \gets X_i(t)$
      \STATE  $\beta = ( 1 +\frac{1}{D_i(t)}) \log (\frac{3 D_i(t) \sqrt{D_i(t)}}{ \delta})$
       \STATE $\gamma = \sup_{s \in [1, D_i(t) - 1]} [s \times \textup{kl}(\tilde{\mu}_{1:s}, \tilde{\mu}_{1:D_i(t)} ) + (D_i(t) - s ) \times \textup{kl}(\tilde{\mu}_{s+1:D_i(t)}, \tilde{\mu}_{1:D_i(t)} )]$
       \IF{ $ \gamma \geq \beta$}
      \STATE $ D_i(t) = 0,  \  \forall \  i \in [N] $ and $\tau \gets t$     
     \ENDIF     
      \ENDFOR
     \STATE  $t \gets t+1$
		\ENDWHILE 
	\end{algorithmic} 
\end{algorithm}
Based on our derivation, GLR-CUCB achieves sublinear growth of AoI Regret with respect to communication round $T$. It should be noted that the GLR-CUCB algorithm outperforms the case where the number of breakpoints \(C_T\) is unknown. This is because when \(C_T\) is known, breakpoint information can be incorporated into the parameter design of the change point detector.

To adapt MAB-based policies to the AoI Bandit setting, we introduce AoI-Aware (AA) variants that incorporate AoI into decision-making. When AoI is below a predefined threshold, the policy follows the original strategy; otherwise, it prioritizes exploitation based on historical observations to mitigate information staleness. Let $\sum a_j(t)$ denote the total AoI across all clients in round $t$, and $\hat{\mu}_i(t)$ represent the estimated transmission success probability of channel $i$ up to round $t$. When AoI is high, the server schedules the $M$ channels with the highest historical success rates to minimize AoI. Otherwise, it focuses on exploring the optimal super-arm via the GLR-CUCB algorithm.

\section{ADAPTIVE CHANNEL MATCHING BASED ON MARGINAL CONTRIBUTION AND FAIRNESS}
Under non-stationary channel conditions, the aggregation state of the federated model becomes inherently unpredictable. From the perspective of performance fairness, the federated model may exhibit a bias toward clients with frequent access to aggregation, thereby degrading the performance of clients assigned to unfavorable channels. Moreover, in terms of collaborative fairness, clients with non-iid distributed data characteristics demonstrate varying marginal contributions, and non-stationary channel states may prevent critical clients from participating in the latest global aggregation. To mitigate these challenges, we propose an adaptive channel matching strategy that incorporates both marginal contribution and collaborative fairness considerations.

Considering a scenario with $N$ sub-channels and $M$ clients, our approach involves scheduling $M$ sub-channels after performing MAB-based channel scheduling in the $t$-th round of aggregation. In piecewise-stationary scenario, we sort each sub-channel according to the UCB value calculated based on Eq. (26), i.e.,
\begin{equation}
 \hat{\mu}_i(t) = \tilde{\mu}_i(t-1) + \sqrt{\frac{3 \log (t - \tau)}{ 2 D_i(t-1) }}.
\end{equation}

For extremely non-stationary channels, in the M-exp3 algorithm we schedule according to the super-arm, i.e., the combination of scheduling channels in each round. Given the absence of independent information specific to individual channels in each round, we rank the $M$ sub-channels according to their historical mean values, i.e.,
\begin{equation}
{{\tilde \mu }_i(t)} = \frac{{\sum\limits_{k = 1}^{t - 1} {{X_i}(} k)}}{{{D_i}(t - 1)}}.
\end{equation}


\begin{figure*}
    \centering
    \includegraphics[width=0.8\linewidth]{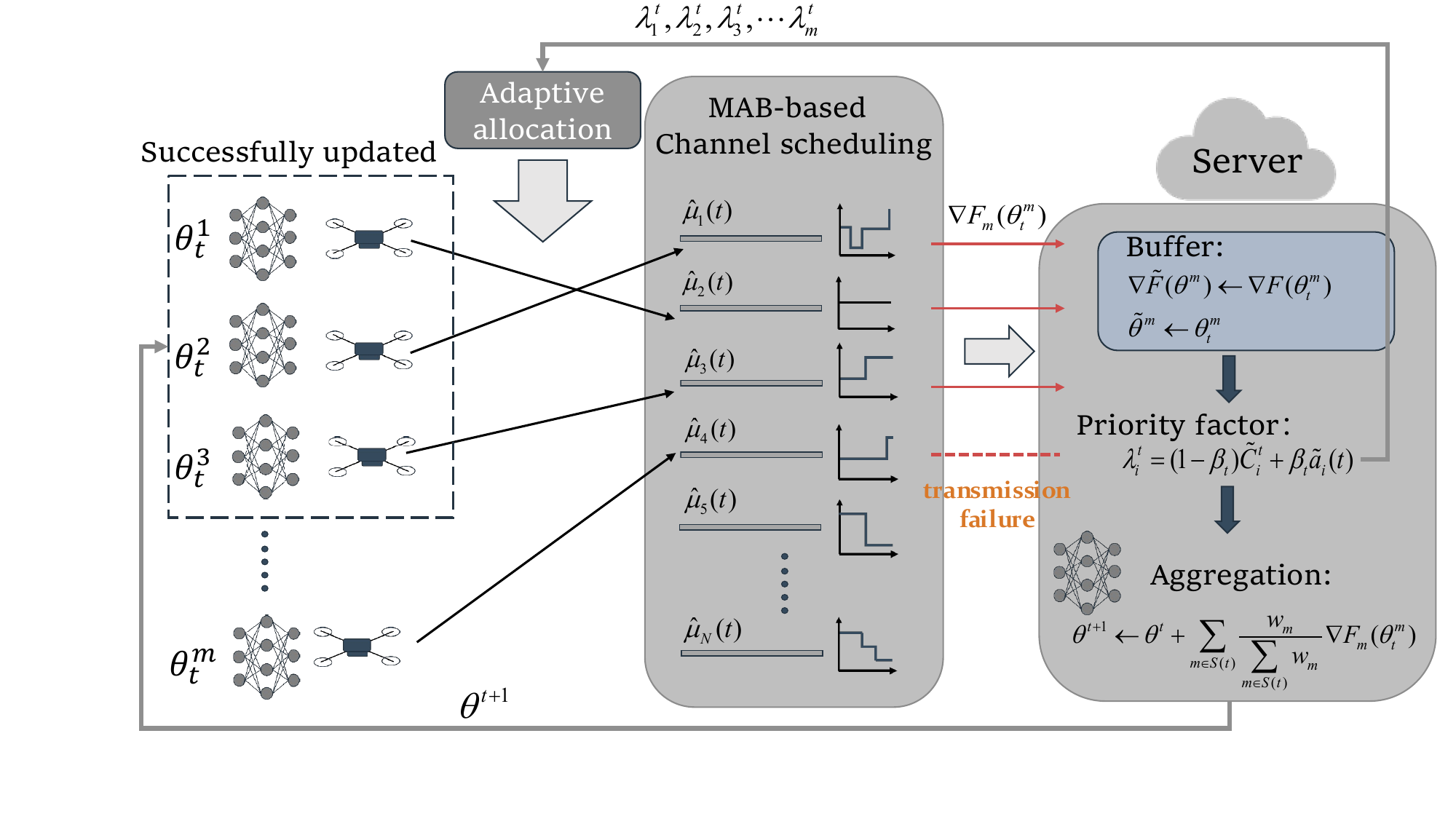}
    \begin{center}
    \caption{Asynchronous federated learning procedure in wireless network}    
    \end{center}
    \label{fig:enter-label}
\end{figure*}
We denote the sorted sub-channels as ${I_1},{I_2},{I_3}, \cdots {I_M}$. Next, we establish a client-channel matching scheme that balances both cooperative fairness and performance fairness. To achieve this, we first define a priority coefficient for each client in a given round, where clients with higher priority are matched to higher-ranked sub-channels. This priority coefficient is dynamically updated based on the client's marginal contribution and AoI. Our proposed matching algorithm is designed to accommodate the adverse effects of channel non-stationarity and adaptively adjust based on feedback from communication outcomes.

The marginal contribution of the client $m$ can be defined in terms of the Shapley value\cite{ghorbani2019data},
\begin{equation}
 {C_m} = \mathbb{E}[U({S_{ - m}} \cup \{ m\} ) - U({S_{ - m}})]  , 
\end{equation}
where $S$ denotes the overall client set and $S_{ - m}$ denotes the set that excluding the client $m$. $U( \cdot )$ is the utility function that measuring the benefit of the client set on global convergence. Due to the computational complexity of directly calculating Shapley values, an estimation method based on marginal contributions that incorporates cosine similarity and model error has been developed \cite{jiang2023fair}. This leads to the following formulation:
\begin{equation}
\tilde{C}_m^t = {\Gamma _{t,m}}(\cos ){\Gamma _{t,m}}(err),
\end{equation}
where
\begin{equation}
    {\Gamma _{t,m}}(\cos ) = 1 - \cos (\nabla {F_m}(\boldsymbol{w} _t^m),\nabla F(\boldsymbol{w} _t^{ - m})),
\end{equation}
\begin{equation}
    \Gamma_{t,m}(err) = \mathcal{E}(\hat{\mathcal{D}}_m; \mathbf{\boldsymbol{w}}_t^{-m}).
\end{equation}


However, in non-stationary channels, clients assigned to poor-quality channels may experience transmission failures, resulting in them lagging behind the global update process. To address this issue, we consider balancing the staleness among clients when allocating channel resources. Specifically, we introduce the variance of AoI to measure the difference in staleness between clients. We have
\begin{equation}
    {{\tilde V}_t} = \frac{{{V_t}}}{{{{\max }_{0 \le \tau  \le t}}\{ {V_t}\} }},
\end{equation}

where

\begin{equation}
    {V_t} = \sum\limits_{i = 1}^M {{{({a_i}(t) - \bar a(t))}^2}},
\end{equation}

\begin{equation}
{{\tilde a}_i}(t) = \frac{{{a_i}(t)}}{{{{\max }_{0 \le \tau  \le t,0 \le j \le m}}\{ {a_j}(\tau )\} }}.
\end{equation}

For client $i$, we define its prioritization factor as
\begin{equation}
\lambda _i^t = (1 - {\beta _t})\tilde C_i^t + {\beta _t}{{\tilde a}_i}(t),
\end{equation}
where
\begin{equation}
    {\beta _t} = \beta {{\tilde V}_t}.
\end{equation}




We sort the clients based on their priority coefficients and match them with the sorted sub-channels. This prioritization facilitates adaptive tuning of channel matching.  





It is worth noting that for lagging clients, the latest local gradient is not available to the server for each communication round. Due to the channel bottleneck, the server is unable to compute the real-time $\Gamma _{t,m}(cos)$ and $\Gamma_{t,m}(err)$ for all clients  $m \in {1,2,\dots, M}$. To mitigate this, we implement a server-side buffer for storing recent gradient information.

For client $m$ in round $t$, the server updates buffer gradient $\nabla {\tilde{{F}}}(\boldsymbol{w} ^m)$ and buffer model parameter ${\boldsymbol{\tilde w}}_m$ with $\nabla F(\boldsymbol{w} _t^{m})$ and $\boldsymbol{w}_t^{ m}$ upon successful parameter transmission. Consequently, we have
\begin{equation}
    \nabla F(\boldsymbol{w} _t^{ - m}) = \frac{{1 - {{\zeta_{m}^{t}}}\nabla {{\tilde F}}({\boldsymbol{w} ^m})}}{{1 - {{\zeta_{m}^{t}}}}},
\end{equation}
and
\begin{equation}
    \boldsymbol{w}_t^{ - m} = \frac{{1 - {\zeta_{m}^{t}}{{\boldsymbol{\tilde w}}_m}}}{{1 - {\zeta_{m}^{t}}}}.
\end{equation}
The server computes $\tilde C_m^t$ via Eq.(29) and updates $\lambda_m^t$ via Eq.(36). Clients are prioritized based on their priority coefficients, with sub-channel $I_i$ allocated to the client with $i$-th highest coefficient. Following local training, clients transmit gradients through assigned channels for global aggregation via Eq.(7), where
\begin{equation}
    {\zeta_{m}^{t}} = \frac{{\tilde C_m^t}}{{\sum\limits_{m = 1}^M {\tilde C_m^t} }}.
\end{equation}

This prioritization facilitates adaptive tuning of channel matching.  When the disparity in staleness among clients is low, indicating relatively equal participation in global aggregation, channel matching prioritizes efficiency. In this scenario, clients with greater marginal benefits can access better channels. Conversely, when the variance of AoI is significant, indicating some clients are notably lagging, channel matching shifts toward balancing staleness and enhancing fairness. Thus, clients with higher AoI are afforded the opportunity to access superior channels. The overall procedure for our proposed scheduling strategy is outlined in Fig.1.

\section{EXPERIMENTS AND NUMERICAL RESULTS}

In this section, we conduct experiments to evaluate the performance of the proposed scheduling algorithms in non-stationary scenarios.

\subsection{Experimental Setup}

We evaluate our proposed approach on the CIFAR-10 and CIFAR-100 image datasets, each containing 60,000 RGB images with 50,000 for training and 10,000 for testing, categorized into 10 and 100 classes, respectively. The model architectures are selected based on task complexity: an eight-layer CNN with 3×3 convolutional layers for CIFAR-10, and ResNet-18 for CIFAR-100.

To simulate non-IID data distributions, we employ the Dirichlet distribution following \cite{li2022federated}. Specifically, we sample $p_k \sim \text{Dir}N(\alpha)$ and allocate a proportion $p_{k,j}$ of class $k$ instances to client $j$, where $\text{Dir}(\cdot)$ denotes the Dirichlet distribution and $\alpha$ is a concentration parameter ($\alpha > 0$). The parameter $\alpha$ controls the degree of data heterogeneity: as $\alpha \to \infty$, the data distribution approaches IID, and $\alpha \to 0$ results in extreme Non-IID conditions.

The experimental setup considers an anarchic and non-stationary channel environment where the central server manages multiple sub-channels without prior knowledge of the time-varying CSI. The channels are vulnerable to potential attacks, leading to state mutations during communication rounds with $C_T$ breakpoints over $T$ rounds. We evaluate our algorithm under two distinct non-stationary scenarios: an extremely dynamic scenario where channel states may change abruptly at any round without warning, and a piecewise-stationary scenario where channel characteristics maintain statistical stationarity between breakpoints. For comparative analysis, we implement a random scheduling baseline where the central server performs random channel scheduling and client-channel matching in each communication round.


All simulation experiments are conducted on an NVIDIA RTX 3090 GPU. The proposed algorithm is comprehensively evaluated against the baseline and its individual components to assess its efficacy in handling non-stationary channel conditions.

\begin{figure}
\centering
\subfloat[Simulation results for the GLR-CUCB and M-exp3]{\includegraphics[width=1.0\linewidth]{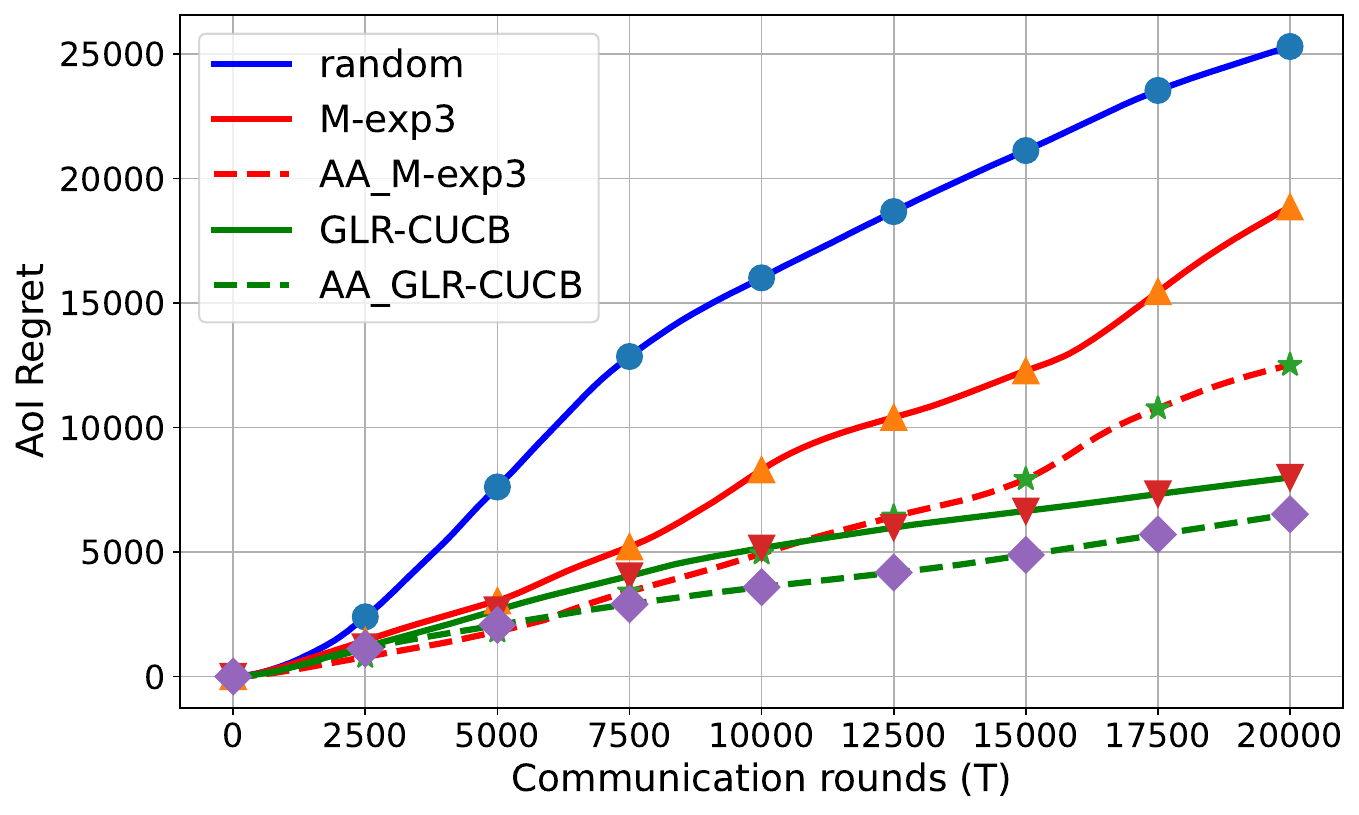}}
\vspace{0.03\linewidth}
\subfloat[Simulation results of M-exp3 with different number of channels]{\includegraphics[width=0.47\linewidth]{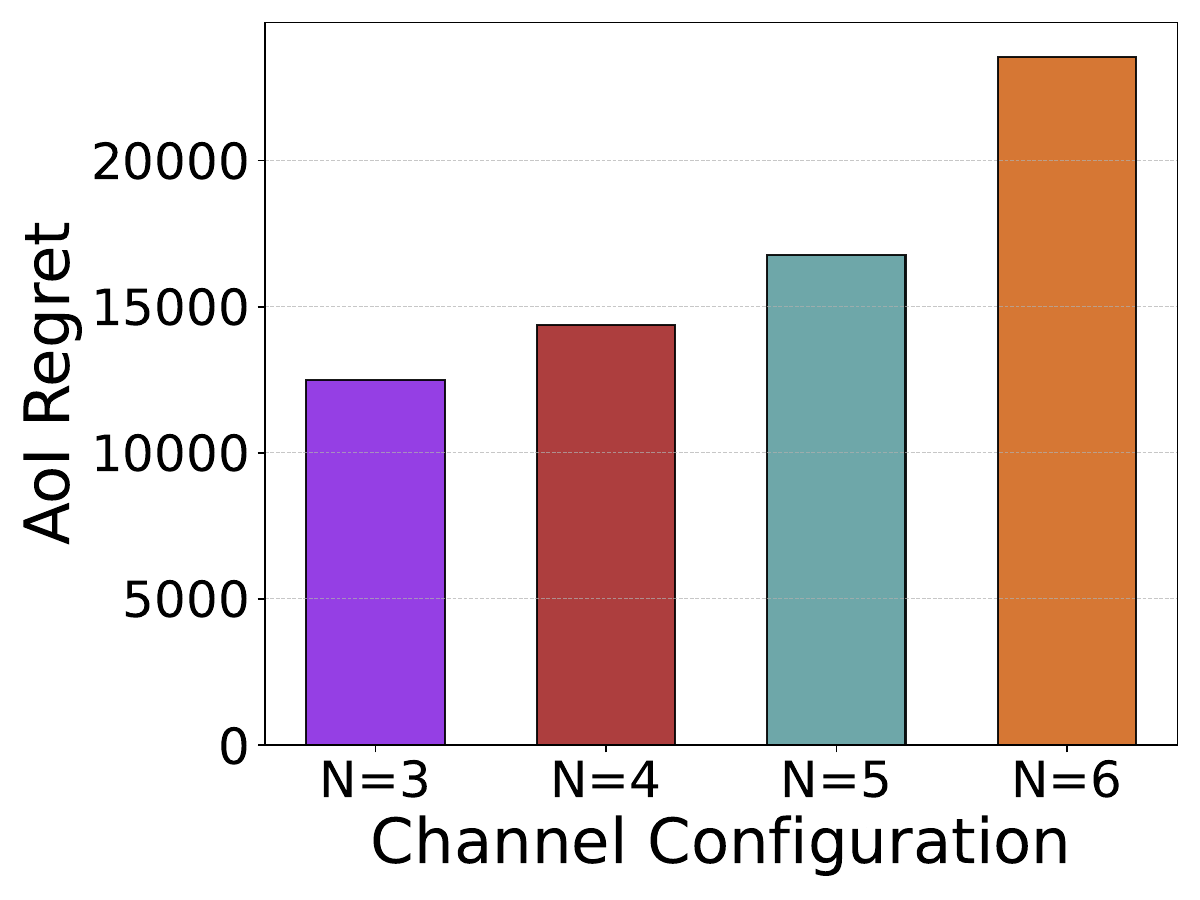}}
\hspace{0.02\linewidth}
\subfloat[Simulation results of GLR-CUCB with different number of breakpoints]{\includegraphics[width=0.47\linewidth]{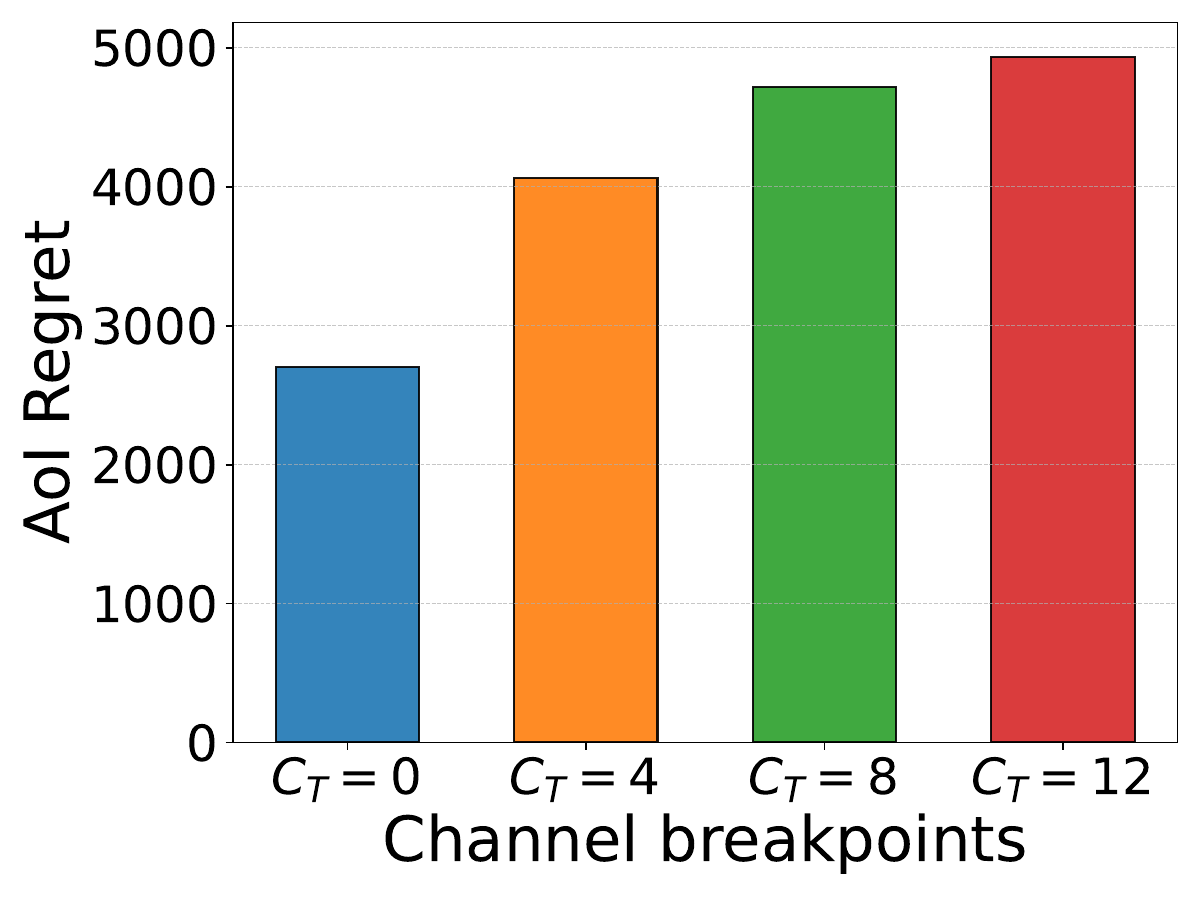}}

\caption{Regret comparison of different algorithms versus communication rounds}
\label{figs}
\end{figure}

\subsection{Performance Comparisons}
\textit{Performance of regret:}
To evaluate the effectiveness of our proposed channel scheduling algorithm in non-stationary environments, we conduct comprehensive ablation experiments. The experimental setup considers channels with $C_T=5$ breakpoints over $T=20000$ rounds, where $M=2$ channels are selected from $N=5$ available channels. The algorithm parameters are configured as follows: $\gamma=0.5$ in Algorithm 1, and $\delta=0.001$, $\alpha=0.05\sqrt{\frac{\log T}{T}}$ in Algorithm 2.
For comparative analysis, we introduce an AoI-Aware (AA) variant that implements a threshold-based scheduling strategy. Specifically, when a client's AoI exceeds the threshold $h(t)$, defined as the inverse of the maximum empirical mean at round $t$, the algorithm directly allocates the channel with the highest historical mean performance to that client.

As demonstrated in Fig.2(a), GLR-CUCB and M-exp3 algorithms exhibit superior performance over random scheduling in identifying high-quality channels and minimizing AoI regret. Both algorithms achieve sub-linear growth in AoI regret, aligning with our theoretical analysis. The integration of AoI awareness enables dynamic channel scheduling based on AoI information, resulting in further reduction of AoI regret. Additionally, GLR-CUCB demonstrates better AoI regret performance compared to M-exp3, consistent with Theorems 3 and 5.

Setting $T=20000$, $M=2$, and $N=5$, Fig. 2(b) illustrates the impact of breakpoints on the GLR-CUCB algorithm's performance. The results demonstrate that the AoI regret consistently increases as the number of breakpoints grows from 0 (representing stationary channels) to 12 within the same number of rounds. An increased number of breakpoints leads to more frequent restarts of the GLR change-point detection mechanism, which in turn affects the algorithm's exploration phase and complicates optimal channel identification.

\begin{figure} 
\centering
\begin{minipage}{\columnwidth} 
    \centering
    \subfloat[Test accuracy on CIFAR-10 in piecewise-stationary channels]{%
        \includegraphics[width=0.8\columnwidth]{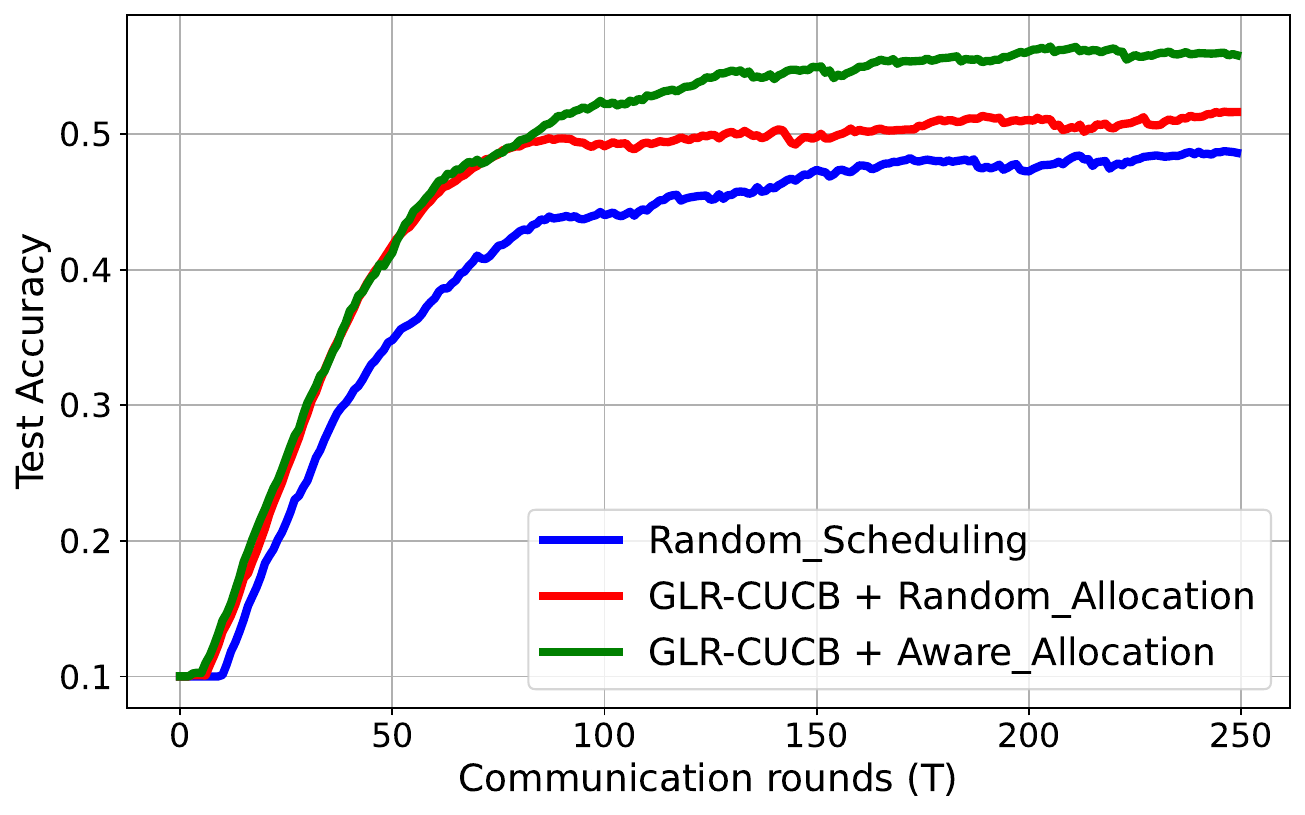}%
        \label{subfig:cifar10_acc}}%
    \vspace{0.3cm} 
    \subfloat[Test accuracy on CIFAR-10 in extremely non-stationary
    channels]{%
        \includegraphics[width=0.8\columnwidth]{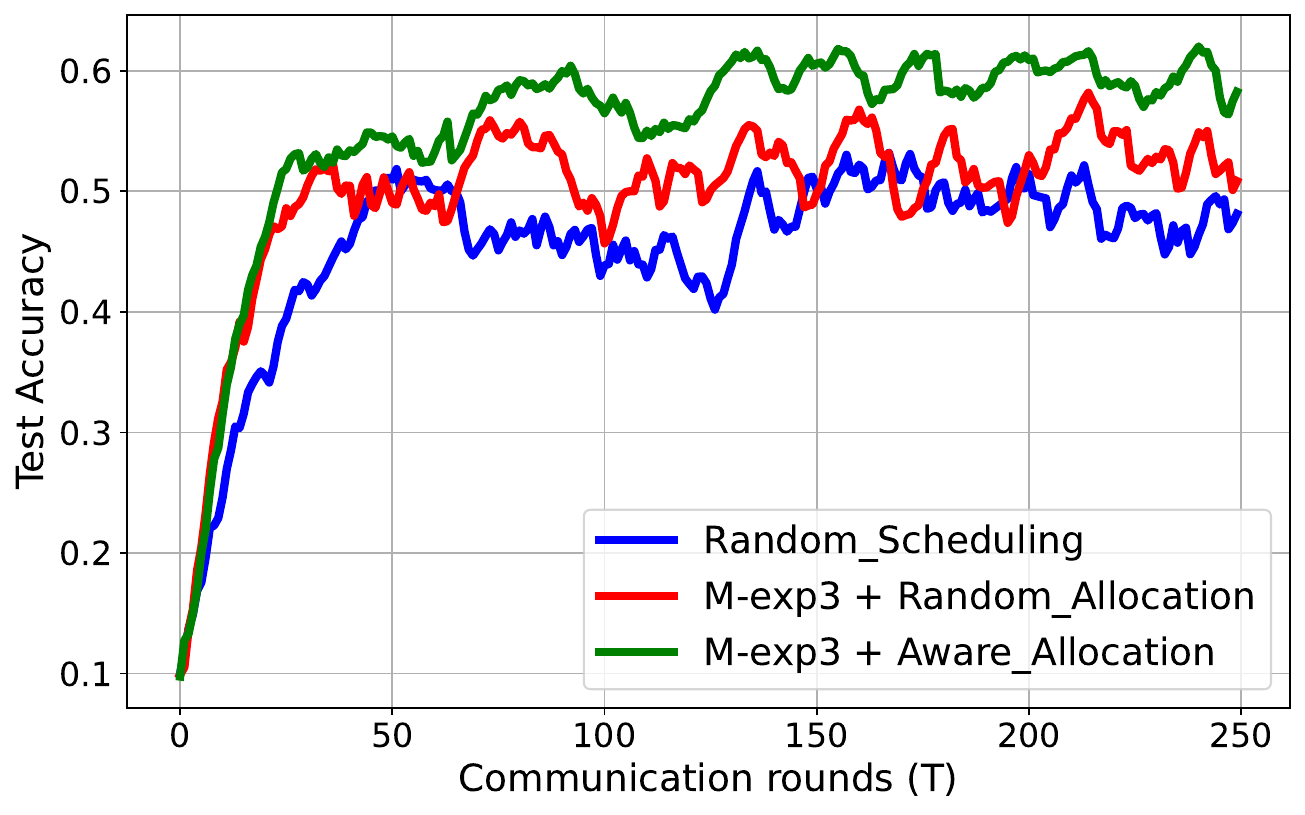}%
        \label{subfig:cifar10_aoi}}%
    \vspace{0.3cm}
    \subfloat[Test accuracy on CIFAR-100 in piecewise-stationary channels]{%
        \includegraphics[width=0.8\columnwidth]{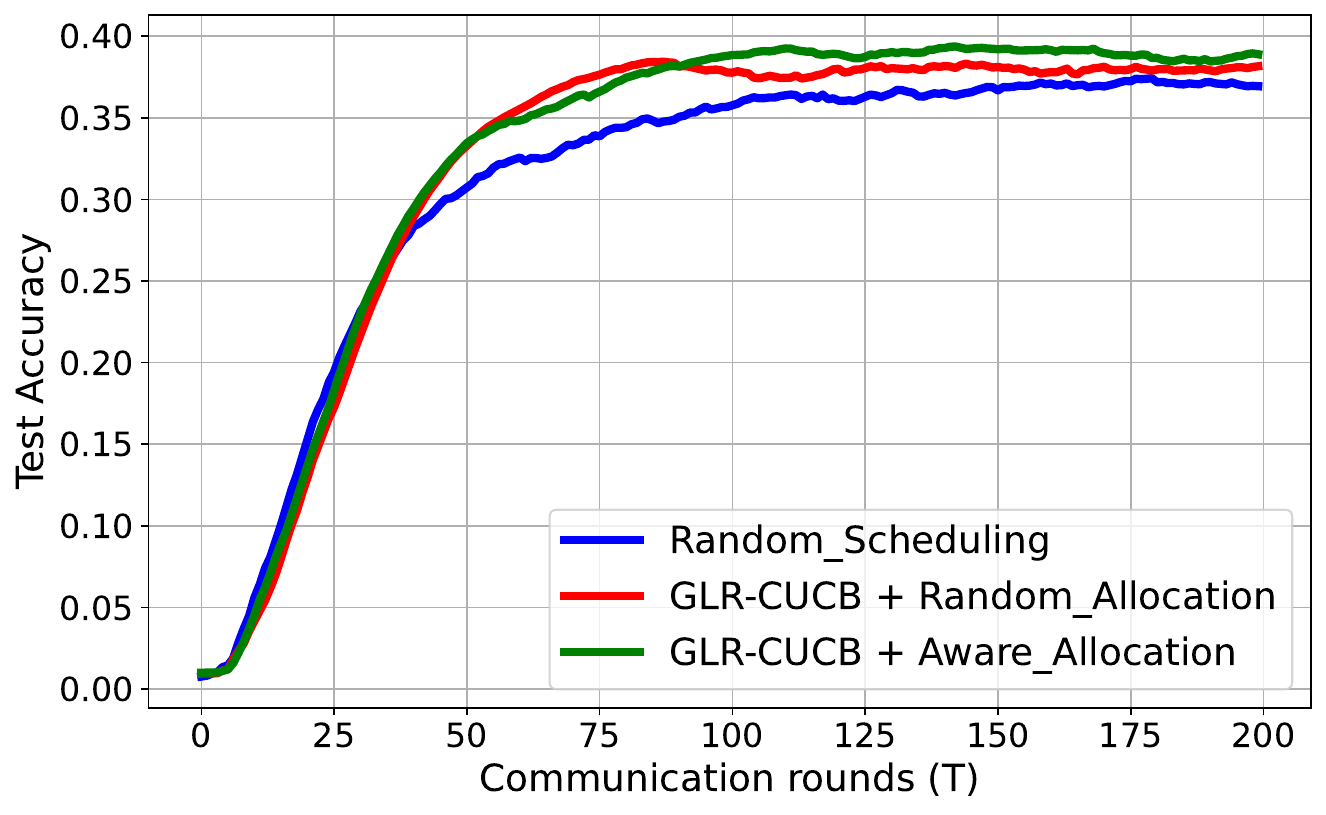}%
        \label{subfig:cifar100_acc}}%
    \vspace{0.3cm}
    \subfloat[Test accuracy on CIFAR-100 in extremely non-stationary
    channels]{%
    \includegraphics[width=0.8\columnwidth]{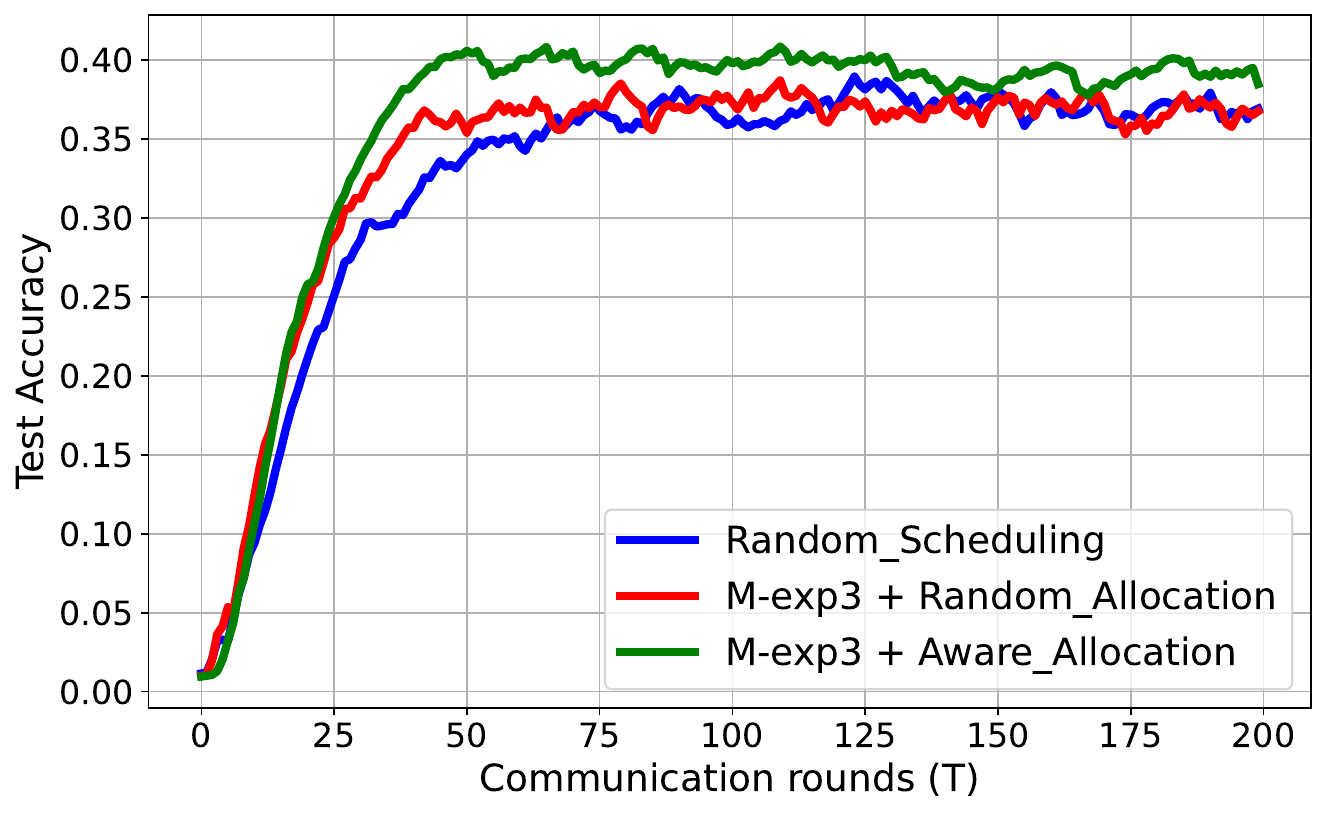}%
        \label{subfig:cifar100_aoi}}%
\end{minipage}
\caption{Performance comparison on test accuracy.}
\label{figs}
\end{figure}


With a fixed number of clients $M=2$, we investigate the impact of varying the number of available channels on system performance. Our findings indicate that increasing the number of available channels does not necessarily enhance performance. For extremely non-stationary channels, the M-exp3 algorithm's performance is constrained by the communication system scale due to the absence of breakpoint information. Fig. 2(c) presents the AoI regret evaluation of M-exp3 across different $C(N,M)$. The results demonstrate that larger $|C(N,M)|$, representing more channel combinations, increase the difficulty for M-exp3 to identify and schedule the optimal super-arm. These experimental observations validate the conclusions drawn in Theorem 2.

\textit{Performance of Test accuracy:} 
We set $\alpha=0.5$ to control the degree of non-iid between clients. To assess the effectiveness of our proposed approach, we conducted ablation experiments examining both the channel scheduling strategy and adaptive channel matching on federated model convergence. The experiments were conducted under two different scales. For piecewise-stationary channel environments, we set $N=30$ channels and $M=20$ clients. Based on our regret analysis results showing the impact of system scale on M-exp3 algorithm performance, we evaluated the extremely non-stationary channel environments in a smaller-scale communication system with $N=6$ channels and $M=4$ clients.




\begin{figure}[htbp]
\centering
\subfloat[AoI variance on CIFAR-10 in piecewise-stationary channels]{\includegraphics[width=0.8\linewidth]{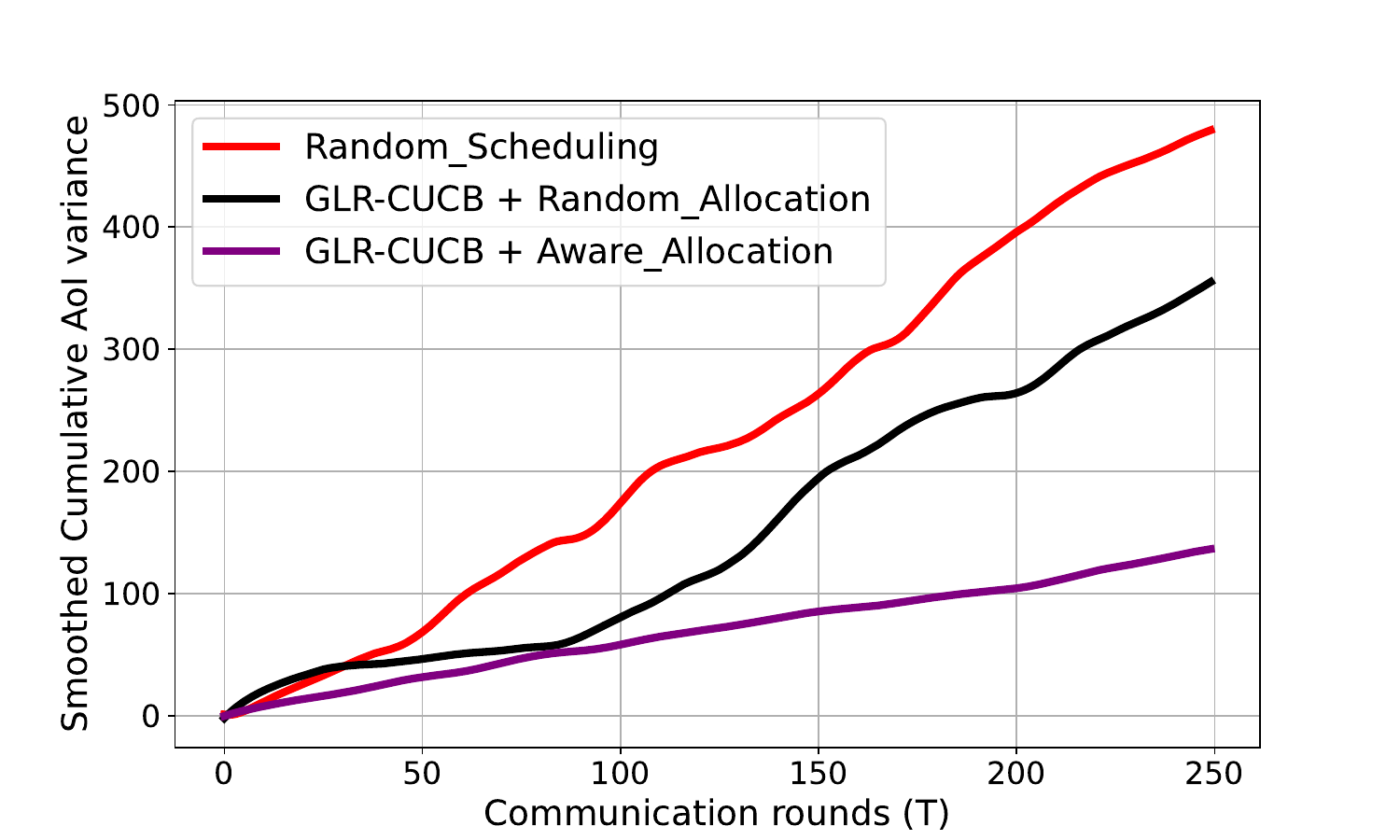}}
\vspace{0.2cm}
\subfloat[AoI variance on CIFAR-10 in extremely non-stationary
    channels]{\includegraphics[width=0.8\linewidth]{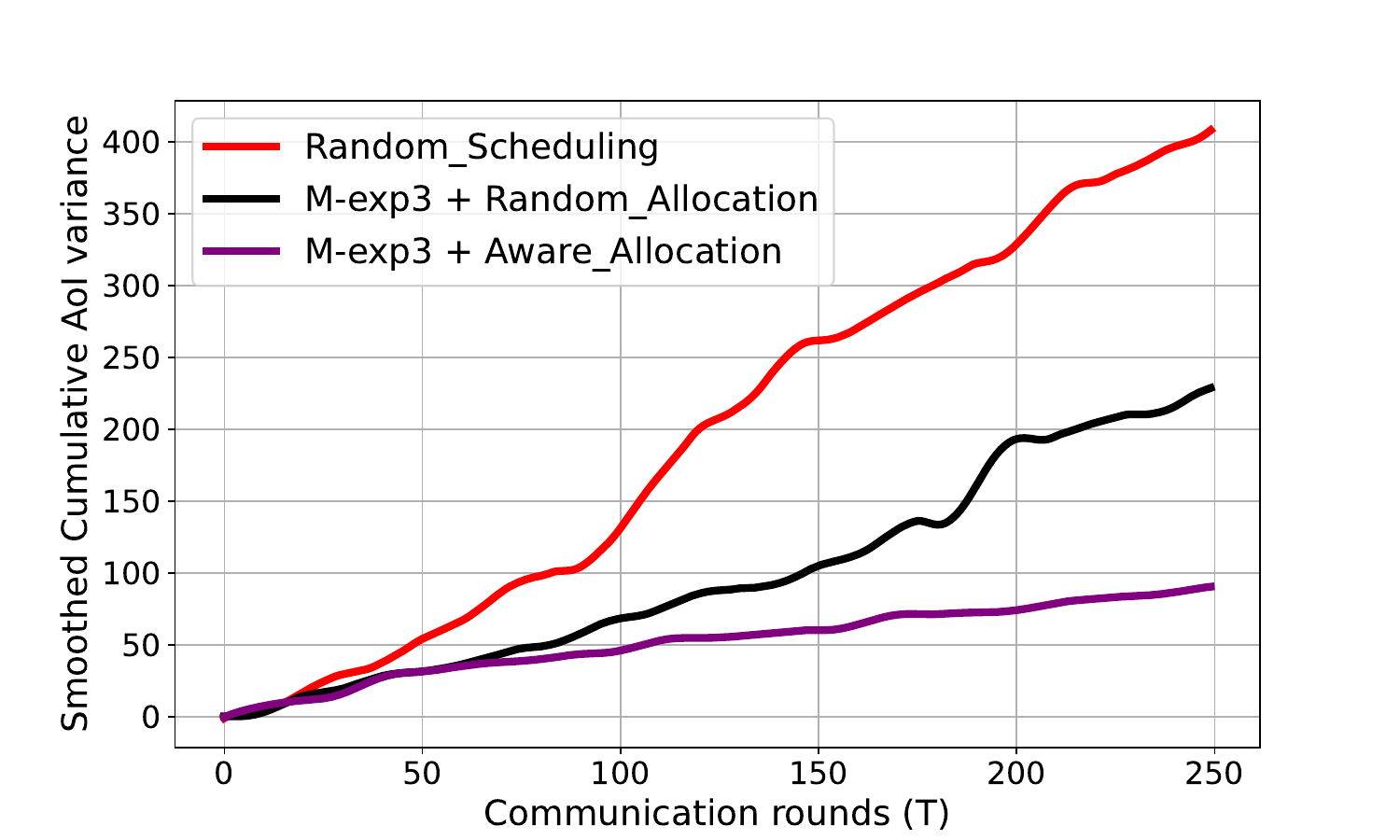}}
\vspace{0.2cm}
\subfloat[AoI variance on CIFAR-100 in piecewise-stationary channels]{\includegraphics[width=0.8\linewidth]{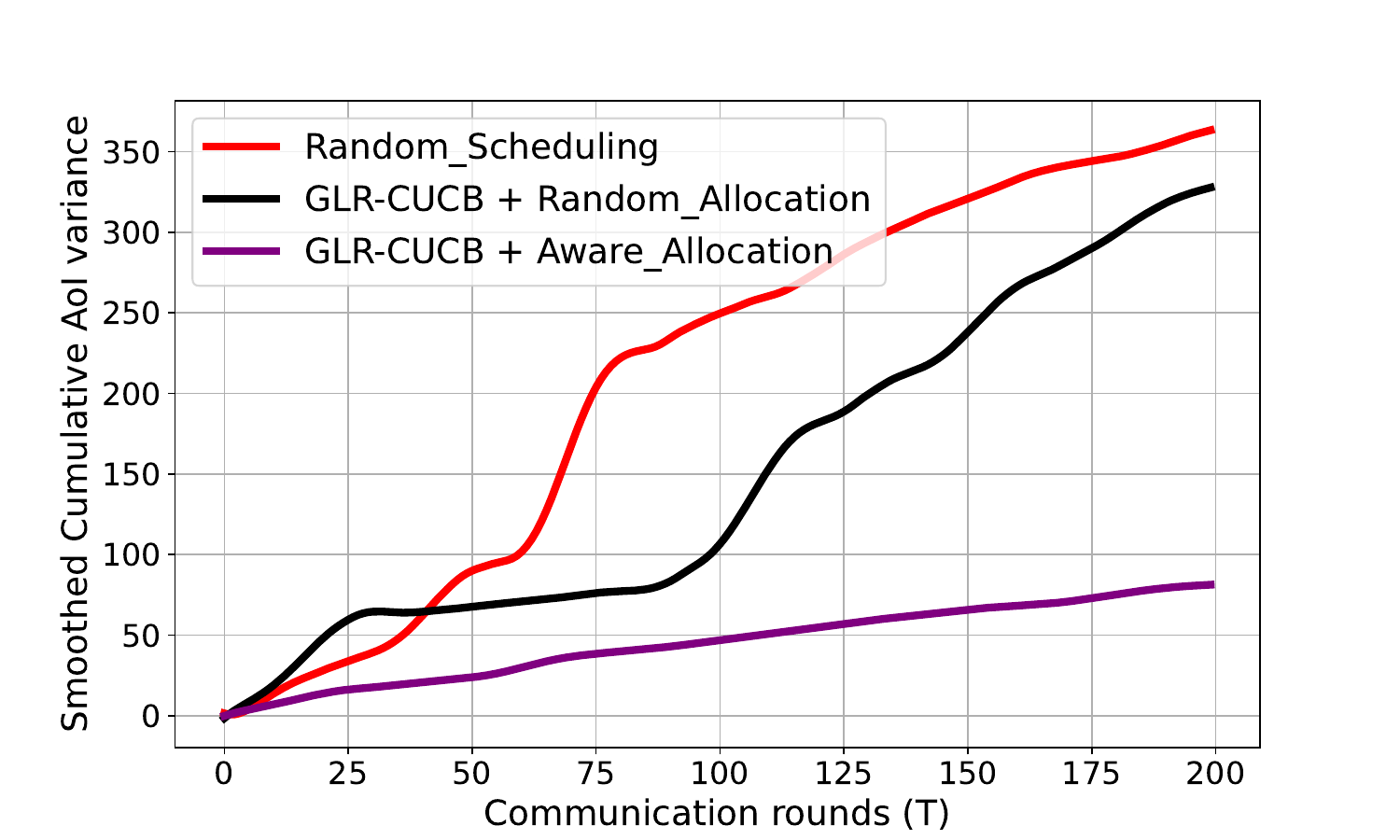}}
\vspace{0.2cm}
\subfloat[AoI variance on CIFAR-100 in extremely non-stationary
    channels]{\includegraphics[width=0.8\linewidth]{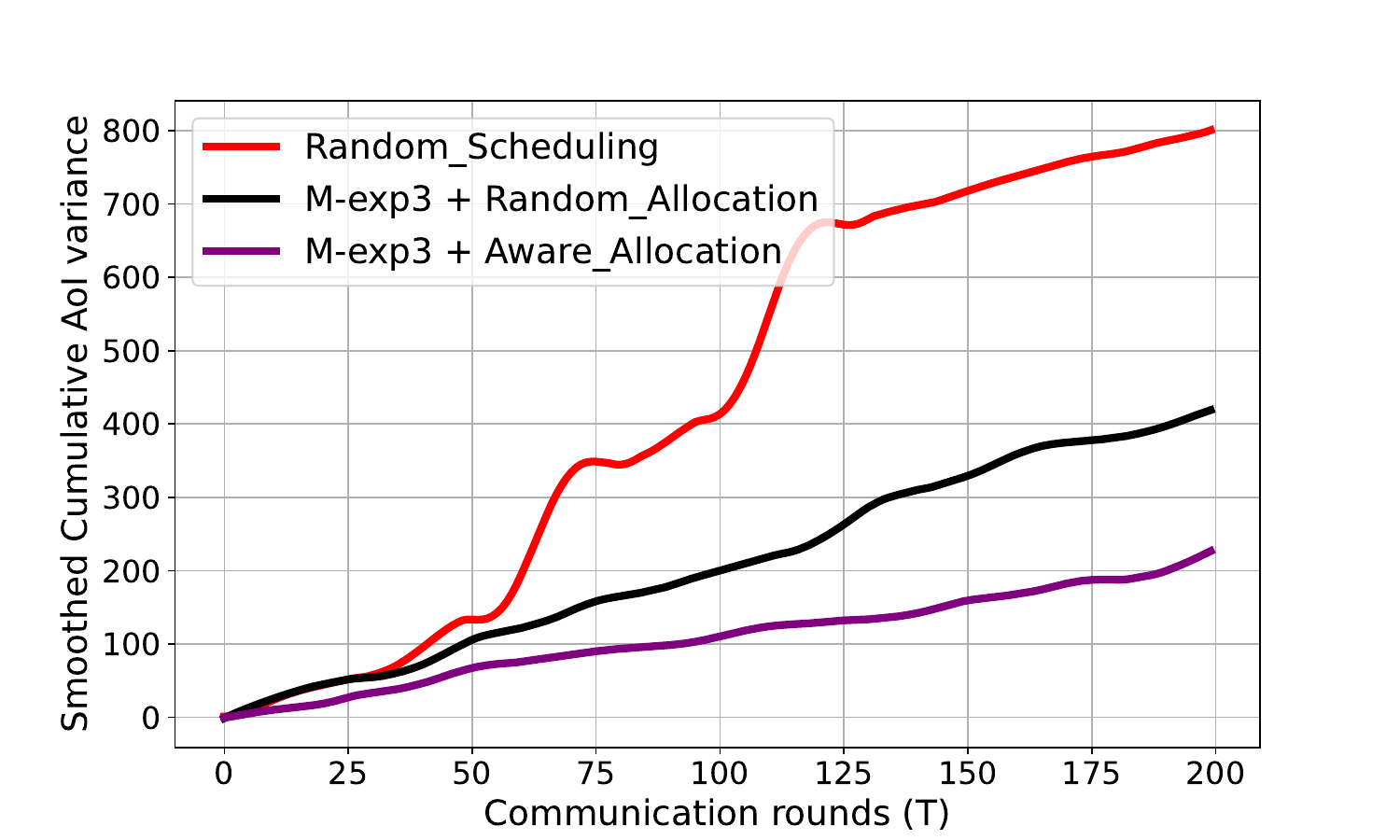}}
\vspace{0.2cm}
\caption{Performance comparison on AoI variance
}
\label{figs}
\end{figure}

As shown in Fig. 3, experimental results on AoI regret validate that GLR-CUCB and M-exp3 enhance communication efficiency and accelerate federated model convergence by mitigating AoI, consistent with our theoretical analysis. In piecewise-stationary channels, GLR-CUCB with aware matching achieves approximately 55\% accuracy on CIFAR-10 and 39\% on CIFAR-100, outperforming random scheduling. Moreover, our proposed algorithms exhibit faster convergence, reaching stable accuracy within 100 communication rounds, compared to 150 rounds for random scheduling. In extremely non-stationary channels, M-exp3 with aware matching maintains a stable performance of approximately 58\% accuracy on CIFAR-10 and 38\% on CIFAR-100, demonstrating improvements over random scheduling and achieving convergence within 75 rounds instead of 100.

The ablation experiments further demonstrate the effectiveness of adaptive channel matching. Comparing the performance between random matching and aware matching, we observe consistent improvements across all scenarios. Specifically, in piecewise-stationary channels, aware matching provides an additional accuracy gain for CIFAR-10 and for CIFAR-100 compared to random matching. This enhancement in performance indicates that the proposed algorithm effectively balances the staleness among clients and mitigates the issue of clients falling significantly behind the global process.

\textit{Performance of fairness:}
As evidenced by the AoI variance results in Figures 4, our proposed algorithms significantly reduce the accumulated AoI variance. In piecewise-stationary channels, GLR-CUCB with aware matching reduces the cumulative AoI variance by approximately 75\% (from 480 to 130) on CIFAR-10 and 77\% (from 360 to 80) on CIFAR-100 compared to random scheduling at round 250. Similarly, in extremely non-stationary channels, M-exp3 with aware matching achieves a reduction of about 75\% (from 400 to 100) on CIFAR-10 and 71\% (from 800 to 230) on CIFAR-100.

From the experimental results, it can be seen that the cumulative AoI variance of the proposed algorithms tends to flatten out with the training process, i.e. the difference in the AoI of the individual clients nearly ceases to change. Specifically, after around 150 communication rounds, the slope of AoI variance curve with aware matching becomes significantly smaller (approximately 0.2) compared to random scheduling (approximately 1.5). We find that the GLR-CUCB as well as M-exp3 algorithms also reduce the AoI variance to some extent, which mainly stems from the ability of the algorithms to reduce the overall AoI of the clients. When certain clients have significantly lagged behind the global update, the adaptive algorithms can dynamically adjust the priority coefficients of clients, allowing the lagging clients to be prioritized for better channels, thereby mitigating client drift. The effectiveness of this approach is particularly evident in the extremely non-stationary scenario, where aware matching maintains the cumulative AoI variance below 100 for CIFAR-10 and 250 for CIFAR-100, compared to random scheduling which reaches 400 and 800 respectively. By balancing the aggregation participation of individual clients, the algorithms facilitate a fairer aggregation process.
\section{Conclusions}
In non-stationary wireless environments, asynchronous federated learning faces stale updates and client drift challenges. We propose a MAB-based channel scheduling with theoretical analysis for two non-stationary scenarios, establishing AoI regret bounds for GLR-CUCB and M-exp3 algorithms. To address imbalanced client updates, we introduce an adaptive channel matching strategy considering marginal utility and fairness. Empirical results demonstrate enhanced communication efficiency with maintained model convergence and fairer client participation in non-stationary federated learning.
\section*{
\begin{center}
    {Appendix A}\\
    {Proof of Theorem 1}
\end{center}
}
For convenience of expression, we denote $\tilde{\eta}=\eta \lambda$, then according to assumption 1, we have

\begin{align}
& F\left(\boldsymbol{w}_{t+1}\right)-F\left(\boldsymbol{w}_t\right) \notag \\
& \leq\left\langle\nabla F\left(\boldsymbol{w}_t\right), \boldsymbol{w}_{t+1}-\boldsymbol{w}_t\right\rangle+\frac{L}{2}\left\|\boldsymbol{w}_{t+1}-\boldsymbol{w}_t\right\|^2 \notag\\
& \leq-\eta\lambda \mathbb{E}\left\langle\nabla F\left(\boldsymbol{w}_t\right), \frac{1}{|S_t| \lambda} \sum_{k\in S_t} \sum_{l=0}^{\lambda-1} \tilde{\nabla} F_k\left(\boldsymbol{w}_{k, t-a_i(t)}^{(l)}\right)\right\rangle  \notag\\
&\hspace{0.5cm}+ \frac{L \eta^2\lambda^2}{2} \mathbb{E}\left\|\frac{1}{|S_t| \lambda} \sum_{i=1}^M \sum_{l=0}^{\lambda-1} \tilde{\nabla} F_k\left(\boldsymbol{w}_{k, t-a_i(t)}^{(l)}\right)\right\|^2 \notag \\
&\leq -\frac{\eta\lambda}{2}\left\|\nabla F\left(\boldsymbol{w}_t\right)\right\|^2 +\underbrace{\frac{\eta\lambda}{2}\mathbb{E}\left\|o_t\right\|^2}_{A_1}  \notag\\
&\hspace{0.5cm}+\underbrace{\frac{L \eta^2\lambda^2}{2} \mathbb{E}\left\|\frac{1}{|S_t| \lambda} \sum_{k\in S_t} \sum_{l=0}^{\lambda-1} \tilde{\nabla} F_k\left(\boldsymbol{w}_{k, t-a_i(t)}^{(l)}\right)\right\|^2}_{A_2},
\end{align}

where $o_t=\frac{1}{|S_t| \lambda} \sum_{k\in S_t} \sum_{l=0}^{\lambda-1} \tilde{\nabla} F_k\left(\boldsymbol{w}_{k, t-a_i(t)}^{(l)}\right)-\frac{1}{M \lambda} \sum_{i=1}^M \sum_{l=0}^{\lambda-1} \tilde{\nabla} F_k\left(\boldsymbol{w}_{k, t-a_i(t)}^{(l)}\right)$, $\tilde{\nabla} F_k(\cdot)$ represents the stochastic gradient and ${\nabla} F_k(\cdot)$ denotes the true gradient. In the following, we bound the terms $A_1$ and $A_2$ respectively. 

To facilitate the derivation, we first introduce the following auxiliary variables:
\begin{equation}
\Gamma_1 = \frac{1}{|S_t| \lambda} \sum_{k\in S_t} \sum_{l  = 0}^{\lambda-1} \tilde{\nabla} F_k\left(\boldsymbol{w}_{k, t-a_i(t)}^{(l)}\right),
\end{equation}
\begin{equation}
\Gamma_2 = \frac{1}{|S_t| \lambda} \sum_{k\in S_t} \sum_{l  = 0}^{\lambda-1} \nabla F_k\left(\boldsymbol{w}_{k, t-a_i(t)}^{(l)}\right),
\end{equation}
\begin{equation}
\Gamma_3 = \frac{1}{M \lambda} \sum_{i=1}^M \sum_{l  = 0}^{\lambda-1} \nabla F_k\left(\boldsymbol{w}_{k, t-a_i(t)}^{(l)}\right),
\end{equation}

\begin{equation}
\Gamma_4 = \frac{1}{M \lambda} \sum_{i=1}^M \sum_{l  = 0}^{\lambda-1} \tilde{\nabla} F_k\left(\boldsymbol{w}_{k, t-a_i(t)}^{(l)}\right).
\end{equation}

Then we have
\begin{equation}
\begin{aligned}
&A_2 = \frac{L \eta^2\lambda^2}{2} \mathbb{E}\left\|\Gamma_1\right\|^2\\
&\leq    \frac{L \eta^2\lambda^2}{2}\{3\mathbb{E}\left\|\Gamma_1-\Gamma_2\right\|^2+3\mathbb{E}\left\|\Gamma_2-\Gamma_3\right\|^2+3\mathbb{E}\left\|\Gamma_3\right\|^2 \}.
\end{aligned}
\end{equation}

It is straightforward to see $\mathbb{E}\left\|\Gamma_1-\Gamma_2\right\|^2 \leq \sigma^2$. For the second and third term in (49), we can derive that
\begin{equation}
\begin{aligned}
&\mathbb{E}\left\|\Gamma_2-\Gamma_3\right\|^2 \\
&  = \left \| \frac{1}{\lambda } \left (\frac{1}{|S_t|} -\frac{1}{M}\right )\sum_{k\in S_t} \sum_{l  = 0}^{\lambda-1} \nabla F_k\left(\boldsymbol{w}_{k, t-a_i(t)}^{(l)}\right)\right.\\
&\hspace{2cm} \left.-\frac{1}{K\lambda }\sum_{k\in \mathcal{K}\setminus {S_t} } \sum_{l  = 0}^{\lambda-1} \nabla F_k\left(\boldsymbol{w}_{k, t-a_i(t)}^{(l)}\right)\right \|^2\\
& \leq \left \|\frac{|S_t|\lambda G}{\lambda } \left (\frac{1}{|S_t|} -\frac{1}{M}\right )+\frac{(K-|S_t|)\lambda G}{K\lambda }\right \|^2\\
&\leq 4\left ( 1-\frac{|S_t|}{K}\right  )^2G^2,  
\end{aligned}
\end{equation}
and
\begin{align}
&\mathbb{E}\left\|\Gamma_3\right\|^2\notag\\
& =\mathbb{E}\left\|\frac{1}{M \lambda} \sum_{i=1}^M \sum_{l=0}^{\lambda-1} \nabla F_k\left(\boldsymbol{w}_{k, t-a_i(t)}^{(l)}\right)\right\|^2\notag \\
& \stackrel{(a)}{\leq} \frac{3}{M \lambda} \sum_{i=1}^M \sum_{l=0}^{\lambda-1} \mathbb{E}\left\|\nabla F_k\left(\boldsymbol{w}_{k, t-a_i(t)}^{(l)}\right)-\nabla F_k\left(\boldsymbol{w}_{t-a_i(t)}\right)\right\|^2\notag \\
& +\frac{3}{K} \sum_{i=1}^M \mathbb{E}\left\|\nabla F_k\left(\boldsymbol{w}_{t-a_i(t)}\right)\!-\!\nabla F_k\left(\boldsymbol{w}_t\right)\right\|^2+3\left\|\nabla F\left(\boldsymbol{w}_t\right)\right\|^2\notag \\
& \leq  \frac{3}{K \lambda} \sum_{i=1}^M \sum_{l=0}^{\lambda-1} L^2 \mathbb{E}\left\|\boldsymbol{w}_{k, t-a_i(t)}^{(l)}-\boldsymbol{w}_{t-a_i(t)}\right\|^2\notag \\
& + \frac{3}{M} \sum_{i=1}^M L^2 \mathbb{E}\left\|\boldsymbol{w}_{t-a_i(t)}-\boldsymbol{w}_t\right\|^2+3\left\|\nabla F\left(\boldsymbol{w}_t\right)\right\|^2,
\end{align}
in which (a) holds because of Jensen's inequality. Substituting the above formula into $A_2$, we have
\begin{equation}
\begin{aligned}
& A_2 \leq \frac{3L \eta^2\lambda^2 \sigma^2}{2}+6L\eta^2\lambda^2\left ( 1-\frac{|S_t|}{K}\right  )^2G^2 \\
&+\frac{9L \eta^2\lambda }{2K } \sum_{i=1}^M \sum_{l=0}^{\lambda-1} L^2 \mathbb{E}\left\|\boldsymbol{w}_{k, t-a_i(t)}^{(l)}-\boldsymbol{w}_{t-a_i(t)}\right\|^2 \\
& + \frac{9 L \eta^2\lambda^2}{2K} \sum_{i=1}^M L^2 \mathbb{E}\left\|\boldsymbol{w}_{t-a_i(t)}\!-\!\boldsymbol{w}_t\right\|^2\!+\!\frac{9L \eta^2\lambda^2}{2}\left\|\nabla F\left(\boldsymbol{w}_t\right)\right\|^2
\end{aligned}
\end{equation}

Similarly, we can bound $A_1$ as follows:
\begin{equation}
\begin{aligned}
&\frac{\eta\lambda}{2}\mathbb{E}\left\|o_t\right\|^2\\
&  =\frac{\eta\lambda}{2} \left \| \frac{1}{\lambda } \left (\frac{1}{|S_t|} -\frac{1}{K}\right )\sum_{k\in S_t} \sum_{l  = 0}^{\lambda-1} \tilde{\nabla} F_k\left(\boldsymbol{w}_{k, t-a_i(t)}^{(l)}\right)\right.\\
&\hspace{0.5cm} \left.-\frac{1}{K\lambda }\sum_{k\in \mathcal{K}\setminus {S_t} } \sum_{l  = 0}^{\lambda-1} \tilde{\nabla} F_k\left(\boldsymbol{w}_{k, t-a_i(t)}^{(l)}\right)\right \|^2\\
& \leq \frac{\eta\lambda}{2}\left \|\frac{|S_t|\lambda G}{\lambda } \left (\frac{1}{|S_t|} -\frac{1}{K}\right )+\frac{(K-|S_t|)\lambda G}{K\lambda }\right \|^2\\
&\leq 2\eta\lambda\left ( 1-\frac{|S_t|}{K}\right  )^2G^2,  
\end{aligned}
\end{equation}
where
\begin{equation}
\begin{array}{l}
\mathbb{E}\left\|\boldsymbol{w}_{m,t}-\boldsymbol{w}_{m,t^{\prime}}\right\|^{2}=\mathbb{E}\left\|\sum_{j=t^{\prime}}^{t-1}\left(\boldsymbol{w}_{j+1}-\boldsymbol{w}_{j}\right)\right\|^{2} \\
=\tilde{\eta}^{2} \mathbb{E}\left\|\sum_{j=t^{\prime}}^{t-1} \frac{1}{\lambda |S_j|} \sum_{k\in S_j}\sum_{l=0}^{\lambda-1} \tilde{\nabla} F_{k}\left(\boldsymbol{w}_{k, j-\tau_{k, j}}^{(l)}\right)\right\|^{2} \\
{\leq} \tilde{\eta}^{2}\left(t-t^{\prime}\right) \sum\limits_{j=t^{\prime}}^{t-1} \mathbb{E}\left\|\frac{1}{\lambda |S_j|} \sum\limits_{k\in S_j} \sum\limits_{l=0}^{\lambda-1} \tilde{\nabla} F_{k}\left(\boldsymbol{w}_{k, j-\tau_{k, j}}^{(l)}\right)\right\|^{2} \\
{\leq} 2 \tilde{\eta}^{2}\left(t-t^{\prime}\right) \sum\limits_{j=t^{\prime}}^{t-1}  \mathbb{E} \| \frac{1}{\lambda |S_j|} \sum\limits_{k\in S_j}\sum\limits_{l=0}^{\lambda-1} \tilde{\nabla} F_{k}\left(\boldsymbol{w}_{k, j-\tau_{k, j}}^{(l)}\right) \\
\hspace{0.5cm} -\nabla F_{k}\left(\boldsymbol{w}_{k, j-\tau_{k, j}}^{(l)}\right ) \|^{2}+ 2 \tilde{\eta}^{2}\left(t-t^{\prime}\right)^2G^2\\
\leq  2 \tilde{\eta}^{2}\left(t-t^{\prime}\right)^2\left (\sigma^2+G^2  \right )
\end{array}
\end{equation}
and
\begin{equation}
\begin{aligned}
&\mathbb{E}\left\|\boldsymbol{w}_{k, t-a_i(t)}^{(l)}-\boldsymbol{w}_{t-a_i(t)}\right\|^2\\
&\leq \left\|\sum_{i=1}^{l-1}(\boldsymbol{w}_{k, t-a_i(t)}^{(i+1)}-\boldsymbol{w}_{k, t-a_i(t)}^{(i)})\right\|^2\\
&\leq l\sum_{i=1}^{l-1}\left\|\boldsymbol{w}_{k, t-a_i(t)}^{(i+1)}-\boldsymbol{w}_{k, t-a_i(t)}^{(i)}\right\|^2,\\
&\leq l^2G^2.
\end{aligned}
\end{equation}

Finally, we have
\begin{equation}
\begin{aligned}
& F\left(\boldsymbol{w}_{t+1}\right)-F\left(\boldsymbol{w}_t\right) \\
&\leq (-\frac{\eta\lambda }{2}+\frac{9L\tilde \eta^2\lambda^2}{2})\left\|\nabla F\left(\boldsymbol{w}_t\right)\right\|^2 \\
&\hspace{0.5cm} +(6L\eta^2\lambda^2+2\eta\lambda)\left ( 1-\frac{|S_t|}{K}\right  )^2G^2\\
&\hspace{0.5cm} +9KL^3\eta ^4\lambda ^4(\sigma ^2+G^2)(\frac{1}{K} \sum_{i=1}^Ma_i(t))^2+A,
\end{aligned}
\end{equation}
where $A=\frac{3L \eta^2\lambda^2 \sigma^2}{2}+\frac{3L^3\eta^2G^2\lambda^3(\lambda-1)(2\lambda -1)}{4}$.
By using the L - Lipschitz continuity of 
$\nabla F\left(\boldsymbol{w}_t\right)$, we have
\begin{equation}
    2L(F(\boldsymbol{w}_t)-F(\boldsymbol{w}^*)) \geq \left\|\nabla F\left(\boldsymbol{w}_t\right)\right\|^2. 
\end{equation}
Substituting (57) into (56), we get the Theorem 1.




\section*{
\begin{center}
    {Appendix B}\\
    {Proof of Theorem 2}
\end{center}
}
This theorem can be proved by modifying the proof of \cite[Theorem 3]{Uchiya2010}. The sequence of Good (1) and Bad (0) channel states can be sampled from the Bernoulli distribution $\mathcal{B}_{\mu}$ where $\mu$ denotes the mean. We consider the arm configuration $\textbf{F}_{(\textbf{i})}$ in which there are only one best super-arm  $\textbf{i} \in C(N,M)$. For any channel $ k \in $ \textbf{i}, its mean is $\mathcal{B}_{\frac{1}{2} + \epsilon}$. For any channel $ k \notin $ \textbf{i}, its mean is $\mathcal{B}_{\frac{1}{2}}$. $C(N,M)$ denotes the set of $M$ arms randomly selected from $N$ arms and $|C(N,M)|$ denotes the size of $C(N,M)$. The AoI regret of every policy is lower bounded by the regret incurred in this bandit problem.

Let $\textbf{k}(t)$ denote the set of $M$ distinct channels choosed in round $t$.  Let $N_{\textbf{i}}$ denote the number of rounds arm $j \in \textbf{i}$ is chosen, namely $N_{\textbf{i}} = \sum_{t = 1}^T |\textbf{k}(t) \cap \textbf{i}|$. For player $i$,  let $k_i(t)$ be the selected arm in round $t$ and  let  $V_i^j(t)$  and $V_i^*(t)$ be indicator random variables denoting
successful transmission in round $t$ on arm $j$ by policy $\pi$ and on
the optimal arm by the oracle policy, respectively.
By definition,
\begin{equation}
\begin{split}
\mathbb{E}[a_i(t)] &= \sum_{\tau=0}^{\infty} \tau \mathbb{P}(a_i(t) = \tau)\\
&= \sum_{\tau=0}^{\infty} \prod \limits_{m=0}^{\tau}(1-\mu_{k_i(t-m)}).
\end{split}
\end{equation}
Since the oracle policy selects the best super arm $\textbf{i}$ and any arm in $\textbf{i}$ is $\mu^*=\frac{1}{2}+\epsilon$ in each round, 
\begin{equation}
\begin{split}
 \mathbb{E}[a_i^{*}(t)] &= \sum_{\tau=0}^{\infty} \prod \limits_{m=0}^{\tau}(1-\mu^{*})=\frac{1}{\mu^*}=\frac{2}{1+2\epsilon}.
\end{split}
\end{equation}
Let $a_i^\pi(t)$ and $a_i^*(t)$ denote the AoI of player $i$ in round $t$ under policy $\pi$ and the oracle policy, respectively. From (8),
\begin{equation}
\begin{split}
a_i^\pi (t) - a_i^*(t) &= (1 - G_i^\pi (t))(a_i^\pi (t - 1) + 1) + G_i^\pi (t) \\
&- (1 - G_i^*(t))(a_i^*(t - 1) + 1) - (G_i^*(t))\\
&\ge (G_i^*(t) - G_i^\pi (t))(a_i^*(t - 1)).\\
\end{split}
\end{equation}
Let $\mathbb{E}_{(\textbf{i})}$ and $\mathbb{P}_{(\textbf{i})}$ denote the expectation and the probability measure  with respect to the arm configuration $\textbf{F}_{(\textbf{i})}$. 
In an alternative coupled system in \cite{krishnasamy2021learning}, $\{ U{(t)_{t \ge 1}}\} $ is a i.i.d. random variables distributed uniformly in (0,1). Let ${R_k}(t) = \mathbb{I}\{U(t) \le {\mu _k}\}$ denote the transmission result on channel $k$ in round $t$, where $\mathbb {I}[ \cdot ]$ denotes the indicator function. 
From (14), (59) and \cite[(5)]{Song2022}, we can get the AoI regret of client $i$ in round $t$.

\begin{align}
R_i^\pi(t; \textbf{F}_{(\textbf{i})}) &= \mathbb{E}_{(\textbf{i})}[a_i^\pi(t) - a_i^*(t)]\notag \\
&\geq  \frac{2}{1 + 2 \epsilon} \mathbb{E}_{(\textbf{i})}[(G_i^*(t)-G_i^\pi(t))] \notag \\
&= \frac{2}{{1 + 2\epsilon}}{\mathbb{E}_{({\bf{i}})}}\left[ {V_i^*(t) - \sum\limits_{k = 1}^N \mathbb{I} \{ {k_i}(t) = k\} V_i^k(t)} \right]\notag \\
&= \frac{2}{1 + 2 \epsilon} \mathbb{E}_{(\textbf{i})}\left[ \sum_{k \notin \textbf{i}} \mathbb{I}\{ k_i(t)=k \} (V_i^*(t)- V_i^k(t)) \right]\notag \\
&= \frac{2}{1 + 2 \epsilon} \sum_{k \notin \textbf{i}}[\mathbb{P}_{(\textbf{i})}(\mathbb{I}\{ k_i(t)=k \}=1) \notag \\
& \qquad  \qquad \qquad  \qquad\times \mathbb{P}_{(\textbf{i})}(\mu_k < U(t) \leq \mu_M^*)]\notag \\
&= \frac{2\epsilon}{1 + 2 \epsilon} \sum_{k \notin \textbf{i}}[\mathbb{P}_{(\textbf{i})}(\mathbb{I}\{ k_i(t)=k \}=1)].
\end{align}

Therefore, we can get the cumulative AoI of all clients in $t$ rounds:
\begin{equation}
\begin{split}
R_{\pi}(T; \textbf{F}_{(\textbf{i})})  &= \sum_{i=1}^{M} \sum_{t = 1}^T R_i^{\pi}(t; \textbf{F}_{(\textbf{i})}) \\
&\geq   \frac{2 \epsilon}{1 + 2 \epsilon} \sum_{i=1}^{M} \sum_{t=1}^T  \left( \sum_{k \notin \textbf{i}}  \mathbb{P}_{(\textbf{i})}(\mathbb{I}\{ k_i(t)=k \}=1) \right) \\
&=\frac{{2\epsilon}}{{1 + 2\epsilon}}\sum\limits_{i = 1}^M {\sum\limits_{k \notin {\bf{i}}} {{\mathbb{E}_{({\bf{i}})}}[D_i^k(T)]} } \\
&= \frac{2 \epsilon}{1 + 2 \epsilon} \mathbb{E}_{(\textbf{i})}[MT - \sum_{t = 1}^T |\textbf{k}(t) \cap \textbf{i}|] \\
&= \frac{2 \epsilon}{1 + 2 \epsilon}  \mathbb{E}_{(\textbf{i})}[MT - N_{\textbf{i}}],
\end{split}
\end{equation}
where $D_i^k(T)$ denotes the number of times client $i$ uses channel $k$ from round $1$ to round $T$.
For an arbitrary policy $\pi$,
\begin{equation}
\begin{split}
R_{\pi}(T) &\geq \max_{\textbf{i} \in C(N,M)} R_{\pi}(T;\textbf{F}_{(\textbf{i})}) \\
& \geq \frac{1}{|C(N,M)|} \sum_{\textbf{i} \in C(N,M)} R_{\pi}(T;\textbf{F}_{(\textbf{i})})\\
& = \frac{2\epsilon}{|C(N,M)|(1+2\epsilon)} \sum_{\textbf{i} \in C(N,M)}\mathbb{E}_{(\textbf{i})}[MT - N_{\textbf{i}}].
\end{split}
\end{equation}
Next, we upper bound $\mathbb{E}_{(\textbf{i})}[N_{\textbf{i}}]$. The key idea is to use an arm configuration $\textbf{F}_{(0)}$ with all arms following $\mathcal{B}_{\frac{1}{2}}$ as the benchmark. From \cite[Theorem 3]{Uchiya2010}, we have
\begin{equation}
\begin{split}
\mathbb{E}_{(\textbf{i})}[N_{\textbf{i}}] \leq \mathbb{E}_{(0)}[N_{\textbf{i}}] + \frac{MT}{2} \sqrt{\mathbb{E}_{(0)}[N_{\textbf{i}}] \log \frac{1}{1-4\epsilon^2}  },
\end{split}
\end{equation}
\begin{equation}
\begin{split}
\sum_{\textbf{i} \in C(N,M)} \mathbb{E}_{(0)}[N_{\textbf{i}}] = |C(N-1, M-1)|MT.
\end{split}
\end{equation}
Then, 
\begin{equation}
\begin{split}
\frac{\sum_{\textbf{i} \in C(N,M)}\mathbb{E}_{(\textbf{i})}[N_{\textbf{i}}]}{|C(N,M)|}  \leq M \left(\frac{MT}{N} + \frac{MT}{2} \sqrt{\frac{T}{N}\log \frac{1}{1-4\epsilon^2}} \right).
\end{split}
\end{equation}
Substitute (66) to (63), we have
\begin{equation}
\begin{split}
R_{\pi}(T) &\geq \frac{2 \epsilon}{1 + 2 \epsilon}M \left(T -  \frac{MT}{N} - \frac{MT}{2} \sqrt{\frac{T}{N}\log \frac{1}{1-4\epsilon^2}}  \right). \\
\end{split}
\end{equation}
We get the Theorem 2 by setting $\epsilon = \frac{1}{4} \frac{N-M}{M \sqrt{NT}}$.

\section*{
\begin{center}
    {Appendix c}\\
    {Proof of Theorem 4}\\
\end{center}
}
\textit{Proof}: We consider a special piecewise-stationary channel distribution  where the best super-arm only changes once. The first $M$ arm is $\mathcal{B}_{\frac{1}{2} + \epsilon}$ where $\frac{1}{2} + \epsilon$ is the means of Bernoulli distribution. The rest arms are $\mathcal{B}_{\frac{1}{2}}$. The best super arm is denoted as $\textbf{f}$. If there are breakpoints, we divide $T$ into $K$ intervals of the same size $\tau$. We denote the channel configuration $\textbf{F}_{(j)}$ that the latter $M$ arms change from $\mathcal{B}_{\frac{1}{2}}$ to $\mathcal{B}_{\frac{1}{2} + \epsilon}$ in the $j$ interval$(1 \le j \le K = [\frac{T}{\tau }])$, and the other arms are $\mathcal{B}_{\frac{1}{2} }$ and $\textbf{I}$ denotes the best super arm. Let $\mathbb{E}_{(j)}$ and $\mathbb{P}_{(j)}$ denote the expectation and probability of $\textbf{F}_{(j)}$.
We denote the channel configuration $\textbf{F}_{(0)}$ that the channel of each arm is $\mathcal{B}_{\frac{1}{2}}$, that is, the stationary channel. Let $\mathbb{E}_{(0)}$ and $\mathbb{P}_{(0)}$ denote the expectation and probability of $\textbf{F}_{(0)}$.

Let $\textbf{k}(t)$ denote the set of $M$ distinct channels choosen at round $t$. Let $N_{\textbf{i}}^{k}$ denote the number of rounds that arm $i \in \textbf{i}=\{\textbf{f},\textbf{I}\}$ is chosen, namely $N_{\textbf{i}}^{k} = \sum_{t = (k-1)\tau+1}^{k\tau} |\textbf{k}(t) \cap \textbf{i}|$. Let $R_{\pi}^k$ denote the AoI regret in the $k$ interval. Since the channel is stationary in each interval and from (62),
\begin{equation}
\begin{split}
R_{\pi}^{k}(\tau) &\geq \frac{2 \epsilon}{1 + 2 \epsilon}  \mathbb{E}[M\tau - N_{\textbf{i}}^{k}].
\end{split}
\end{equation}

For the channel configuration $\textbf{F}_{(j)}$,

\begin{equation}
\begin{split}
R_{\pi}(T; \textbf{F}_{(j)})  &= \sum_{k=1}^{K} R_{\pi}^{k}(\tau)\\
&=\sum_{k\neq j}R_{\pi}^{k}(\tau)+R_{\pi}^{j}(\tau)\\
&\geq \frac{2 \epsilon}{1 + 2 \epsilon}  \mathbb{E}_{(j)}[M\tau - N_{\textbf{I}}^{j}]. \\
\end{split}
\end{equation}

Therefore,
\begin{equation}
\begin{split}
R_{\pi}(T) &\geq \max_{j=1,\cdots,K}R_{\pi}(T; \textbf{F}_{(j)}) \\
&\geq \frac{1}{K} \sum_{j=1}^{K}R_{\pi}(T; \textbf{F}_{(j)})  \\
&=\frac{2 \epsilon}{1 + 2 \epsilon} \left( M\tau- \frac{1}{K}\sum_{j=1}^{K}\textbf{E}_{(j)}[N_{\textbf{I}}^{j}]\right).\\
\end{split}
\end{equation}

From (64), 
\begin{equation}
\begin{split}
\mathbb{E}_{(j)}[N_{\textbf{L}}^{j}] \leq \mathbb{E}_{(\textbf{0})}[N_{\textbf{I}}^{j}]+ \frac{M\tau}{2} \sqrt{\mathbb{E}_{(\textbf{0})}[N_{\textbf{I}}^{j}] \log \frac{1}{1-4\epsilon^2}  }.
\end{split}
\end{equation}
Therefore,

\begin{equation}
\begin{split}
&\sum_{j=1}^{K}\mathbb{E}_{(j)}[N_{\textbf{I}}^{j}] \\
&\leq \sum_{j=1}^{K} \mathbb{E}_{(\textbf{0})}[N_{\textbf{I}}^{j}]+ \frac{M\tau}{2} \sqrt{\sum_{j=1}^{K} \mathbb{E}_{(\textbf{0})}[N_{\textbf{I}}^{j}] \log \frac{1}{1-4\epsilon^2}}\\
&=\mathbb{E}_{(\textbf{0})}[N_{\textbf{I}}]+\frac{M\tau}{2} \sqrt{\sum_{j=1}^{K} \mathbb{E}_{(\textbf{0})}[N_{\textbf{I}}] \log \frac{1}{1-4\epsilon^2}  }.
\end{split}
\end{equation}

For $T$, if $\mathbb{E}_{(\textbf{0})}[N_{\textbf{I}}] \leq M\tau$, then
\begin{equation}
\begin{split}
R_{\pi}(T)\geq \frac{2\epsilon}{1+2\epsilon} \left( M\tau -\frac{M\tau}{K}-\frac{M\tau}{2K} \sqrt{M\tau \log \frac{1}{1-4\epsilon^2}}\right).
\end{split}
\end{equation}
From $\tau=\frac{T}{K}$, let $K=N$,
\begin{equation}
\begin{split}
R_{\pi}(T)\geq \frac{2\epsilon}{1+2\epsilon} \left( \frac{MT}{N} -\frac{MT}{N^2}-\frac{MT}{2N^2} \sqrt{\frac{MT}{N} \log \frac{1}{1-4\epsilon^2}}\right).
\end{split}
\end{equation}
Let $\epsilon=\sqrt{\frac{N}{MT}}$, we arrive at Theorem 4.

\bibliography{v4}

\end{document}